\documentclass[lettersize,journal]{IEEEtran}
\usepackage{amsmath,amsfonts}
\usepackage{algorithm}
\usepackage{array}
\usepackage[caption=false,font=normalsize,labelfont=sf,textfont=sf]{subfig}
\usepackage{textcomp}
\usepackage{stfloats}
\usepackage{url}
\usepackage{verbatim}
\usepackage{graphicx}
\usepackage{algpseudocode}
\usepackage[numbers]{natbib}
\usepackage{multirow}
\usepackage{amssymb} 
\usepackage[colorlinks,
            linkcolor=black,
            anchorcolor=white,
            citecolor=black]{hyperref}

\begin{document}

\title{Information-containing Adversarial Perturbation  for Combating Facial Manipulation Systems}

\author{Yao Zhu, Yuefeng Chen, Xiaodan Li, Rong Zhang, Xiang Tian, Bolun Zheng, Yaowu Chen
\thanks{Yao Zhu is with the College of Biomedical Engineering $\&$ Instrument Science, Zhejiang University, Hangzhou 310027, China (E-mail: ee$\_$zhuy$@$zju.edu.cn).}
\thanks{Yuefeng Chen, Xiaodan Li and Rong Zhang are with the Security Department of Alibaba Group, Hangzhou 311121, China (E-mail: 
yuefeng.chenyf@alibaba-inc.com,fiona.lxd@alibaba-inc.com, stone.zhangr@alibaba-inc.com).}
\thanks{Xiang Tian is with the College of Biomedical Engineering $\&$ Instrument Science, Zhejiang University, Hangzhou 310027, China, and also with Zhejiang Provincial Key Laboratory for Network Multimedia Technologies, Hangzhou 310027, China (E-mail: tianx@zju.edu.cn).}
\thanks{Bolun Zheng is with the School of Automation, Hangzhou Dianzi University, Hangzhou 310018, China, and also with Zhejiang Provincial Key Laboratory for Network Multimedia Technologies, Hangzhou 310027, China (E-mail: blzheng@hdu.edu.cn).}
\thanks{Yaowu Chen is with the College of Biomedical Engineering $\&$ Instrument Science, Zhejiang University, Hangzhou 310027, China, and also with Zhejiang University Embedded System Engineering Research Center, Ministry of Education of China, Hangzhou 310027, China (E-mail: cyw@mail.bme.zju.edu.cn).}
\thanks{Corresponding authors: Xiang Tian, Bolun Zheng.}

}



\maketitle

\begin{abstract}
  With the development of deep learning technology, the facial manipulation system has become powerful and easy to use. Such systems can modify the attributes of the given facial images, such as hair color, gender, and age. Malicious applications of such systems pose a serious threat to individuals' privacy and reputation. Existing studies have proposed various approaches to protect images against facial manipulations. 
  Passive defense methods aim to detect whether the face is real or fake, which works for posterior forensics but can not prevent malicious manipulation.
  {Initiative defense methods protect images upfront by injecting adversarial perturbations into images to disrupt facial manipulation systems but can not identify whether the image is fake.}
 
 To address the limitation of existing methods, we propose a novel two-tier protection method named \textbf{I}nformation-containing \textbf{A}dversarial \textbf{P}erturbation (\textbf{IAP}), which provides more comprehensive protection for {facial images}. We use an encoder to map a facial image and its identity message to a cross-model adversarial example which can disrupt multiple facial manipulation systems to achieve initiative protection. Recovering the message in adversarial examples with a decoder serves passive protection, contributing to provenance tracking and fake image detection.
 We introduce a feature-level correlation measurement that is more suitable to measure the difference between the facial images than the commonly used mean squared error.
 Moreover, we propose a spectral diffusion method to spread messages to different frequency channels, thereby improving the robustness of the message against facial manipulation. 
 Extensive experimental results demonstrate that our proposed \textbf{IAP} can recover the messages from the adversarial examples with high average accuracy and effectively disrupt the facial manipulation systems.

\end{abstract}

\begin{IEEEkeywords}
Adversarial attack, adversarial perturbation, face protection, facial manipulation.
\end{IEEEkeywords}

\section{Introduction}

\IEEEPARstart{D}{eep} neural networks (DNNs) have achieved impressive success in many fields. However, this technology can be easily misused and maliciously damaged, which may threaten security or human privacy. On the one hand, \citet{szegedy2013intriguing} found that classification models are vulnerable to imperceptible adversarial perturbation in images, and \citet{zhu2022rethinking} show that attackers can perform black-box attacks which have no access to the target classifier with a high success rate. The existence of adversarial examples raises concerns about the deployment of deep neural networks in real-world scenarios, such as face recognition~\cite{TIP_face_attack,facerec_tifs3,facerec_tifs2} and autonomous driving~\cite{autonomous_tifs,autonomous_cvprwork2,autonomous_cvprwork}. {On the other hand, the advancement of image translation techniques based on generative models has laid the groundwork for implementing realistic facial processing systems\cite{guera2018deepfake,meskys2020regulating}.} Currently, some publicly available applications can realize face editing \cite{thies2016face2face, coleman2019deepfake}, face reenactment \cite{yao2020mesh} and face swapping \cite{korshunova2017fast}. Abuse of these technologies will result in numerous legal, social, and ethical issues.

\begin{figure}[htbp!]
\centering
\includegraphics[width=0.495\textwidth]{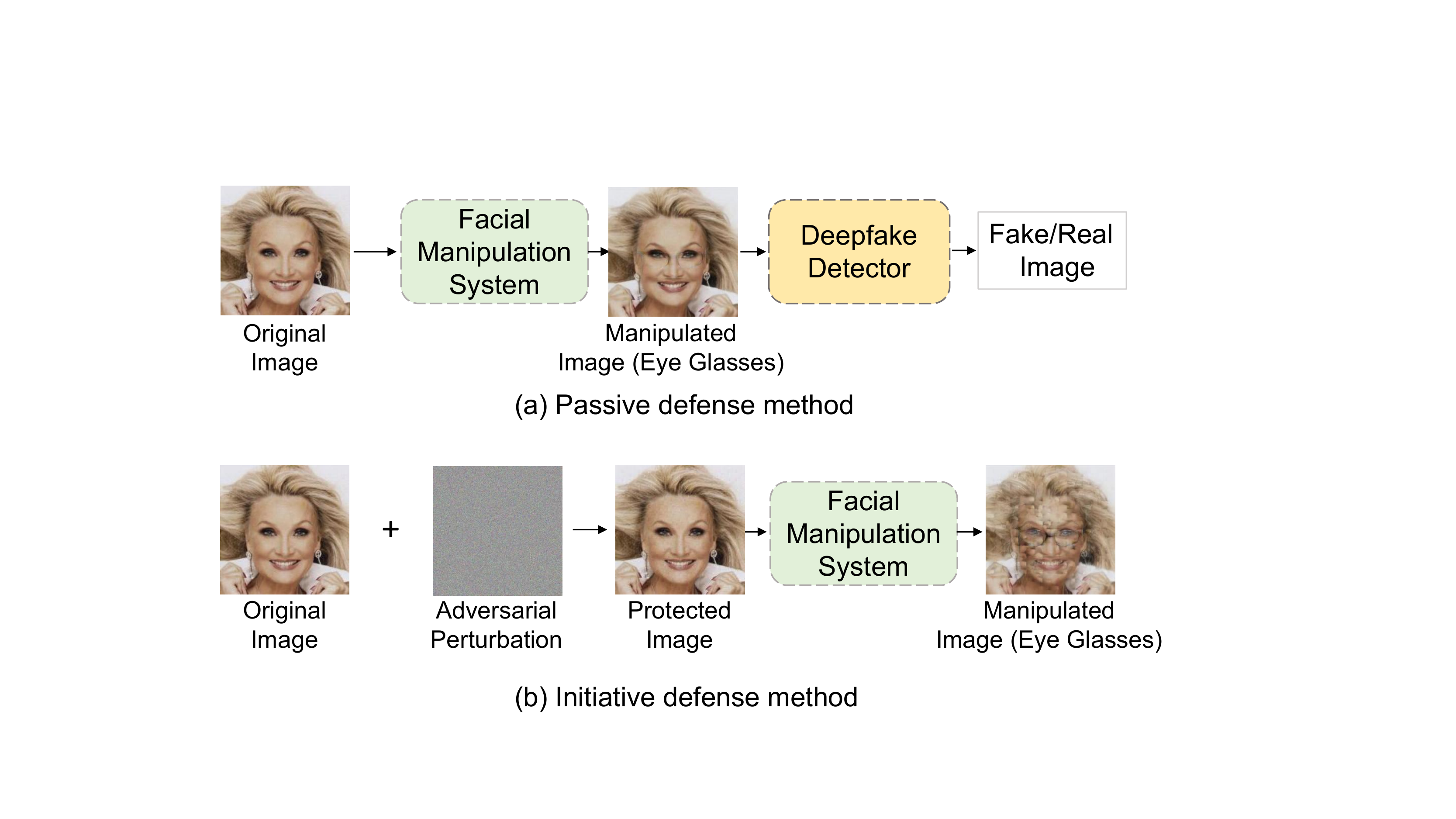}
\caption{Passive defense (a) uses a Deepfake detector to detect whether the image is fake, which can only posteriorly defend {malicious manipulation}. Initiative defense (b) uses adversarial perturbation to disrupt the facial manipulation systems to generate perceptibly distorted outputs, which reduce the risk in advance.}
\label{img:Initiative_Passitive}
\end{figure}

In this paper, we mainly concentrate on the security concerns associated with protecting facial images from malicious manipulation. Existing methods can be roughly divided into the passive defense method and the initiative defense method. Fig. \ref{img:Initiative_Passitive} illustrates the schematic comparison of these two methods. Passive defense methods \cite{dang2020detection,li2018ictu,fernandes2019predicting} train the Deepfake detectors, which are essentially binary classifiers, to {identify} fake images by detecting the artifacts caused by facial manipulation systems. This kind of method passively detects whether malicious manipulation has occurred and can not prevent malicious manipulation from happening. Initiative defense methods \cite{ruiz2020disrupting,huang2021initiative,huang2021cmuawatermark} inject imperceptible adversarial perturbation into the image to protect the image, which can disturb the generative capability of facial manipulation systems. The adversarial perturbation can be added to facial images in advance so that they can avert the risk of malicious manipulation as an initiative defense. However, the {standard adversarial} perturbation usually seems like messy noise that is not conducive to further analysis of the image.
For instance, {the standard adversarial perturbation can not indicate the protection states of the image and can not be used to trace the source of the image}.

\begin{figure*}[t]
\centering
\includegraphics[width=0.97\textwidth]{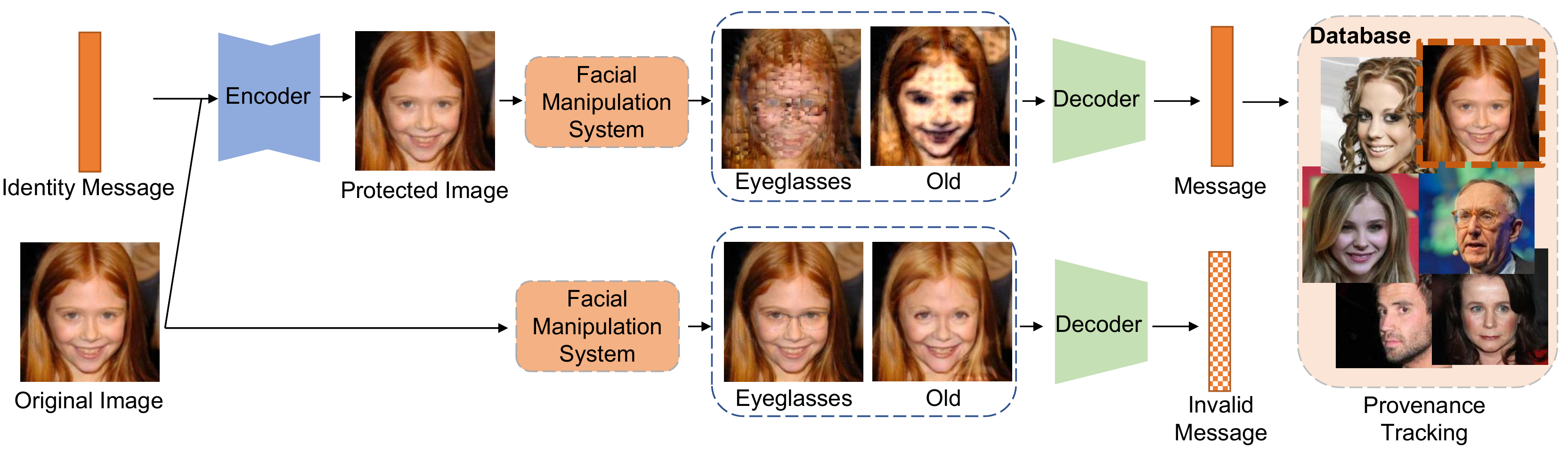}
\caption{Illustration of our \textbf{I}nformation-containing \textbf{A}dversarial \textbf{P}erturbation (\textbf{IAP}). We use an encoder to map a facial image and its corresponding identity message into an adversarial example (the protected image). This adversarial example is visually similar to the original image but can disrupt the facial manipulation system. The message contained in it persists after facial manipulation, which can be used to trace the source of the image in the database. The message extracted from the protected images is close to 0 or 1 at each bit while the message extracted from the unprotected images is close to 0.5 (See more details in section \ref{ablation}). {We regard the message close to 0.5 at each bit as invalid.}}
\label{img:intro}
\end{figure*}

To address the limitations of existing methods, we propose using \textbf{i}nformation-containing \textbf{a}dversarial \textbf{p}erturbation (\textbf{IAP}) to provide two-tier protection for facial images spread in the social media platform. 
We construct a database with the facial image with a unique index, and we term this unique index as the identity message for the image. 
As illustrated in Fig.~\ref{img:intro}, we use an encoder to map the facial image and its identity message to an adversarial example and recover the message contained in the image with a decoder.
{On the one hand, \textbf{IAP} is an adversarial perturbation, which can distort the output of multiple facial manipulation systems to achieve initiative defense. 
On the other hand, the message extracted from the protected image can be used as a clue for tracing the source of the image in the database. Thus we can detect whether the image is manipulated by comparing it with the source image in the database to achieve passive defense.} Moreover, the message extracted from the protected image is close to 0 or 1, while the message extracted from the unprotected image is close to 0.5 (See section \ref{ablation} for details). Therefore, the extracted message can indicate the protection state of the image.

 {
 In designing our IAP, we should address the following three focal challenges.} \textbf{1) The effectiveness of the adversarial perturbation.}  We analyze the defects of the existing loss function used to generate adversarial perturbations against the facial manipulation systems and propose a feature-level correlation measurement method, which can better measure the correlation between damaged and original facial images and adapt well to the multi-optimization problem in training the encoder and decoder. \textbf{2) The correctness of recovering the message contained in the adversarial example.} Facial manipulation can discard some information contained in the adversarial examples while editing the face and make it challenging to recover the information from maliciously manipulated images. We observe that malicious manipulation greatly impacts the frequency domain of the image, so we propose a frequency diffusion method to diffuse the message in different frequency channels of the image, thereby alleviating the effects of damaged frequency channels. \textbf{3) The visual quality of the adversarial image.} We use a discriminator to work as an adversary against the encoder to encourage the encoder to generate more realistic images, which is inspired by the generative adversarial networks (GANs) \cite{goodfellow2014generative,huang2020pfa}. Experimental results show that \textbf{IAP} has achieved good performance in disrupting facial manipulation systems and recovering the messages contained in adversarial examples while maintaining the visual quality of facial images.
 
 The main contributions of our paper are summarized as follows:
 \begin{itemize}
 \item[$\bullet$] We propose a novel two-tier protection method named \textbf{i}nformation-containing \textbf{a}dversarial \textbf{p}erturbation (\textbf{IAP}) for protecting facial images' privacy. \textbf{IAP} initiatively disrupts the facial manipulation systems and can passively trace the source of the manipulated image.
 \item[$\bullet$] We analyze the deficiencies of existing pixel-level loss functions for attacking facial manipulation systems and propose a feature-level correlation measurement method to address these deficiencies. We also propose a frequency diffusion method to spread the information contained in \textbf{IAP} to different frequency channels to improve the robustness of information against different facial manipulation.
 \item[$\bullet$] Experimental results demonstrate that our \textbf{IAP} can defend against malicious behaviors of multiple facial manipulation systems and extract information from adversarial examples correctly, providing a high-quality solution for face privacy protection. 
 \end{itemize}
 
The rest of this paper is organized as follows. {Section II provides some preliminaries for the {standard adversarial} perturbation, the difference between our information-containing adversarial perturbation and the {standard adversarial} perturbation, and shows how to use our IAP to protect images spread in social media platforms.} Section III summarizes the literature on deep face forgery and the defense methods against facial manipulation. 
In Section IV, we first show the overview of the architecture of IAP. Then we introduce each module in detail. We analyze the limitations of the currently used MSE loss functions for attacking facial manipulation systems and propose our feature-level correlation measurement. After that, we propose a spectral diffusion module to improve the robustness of the information contained in adversarial examples against different facial manipulation systems.
In Section V, we first conduct experiments to demonstrate the superiority of \textbf{IAP} in disrupting facial manipulation systems. Then, we evaluate the robustness of the information contained in the \textbf{IAP} against malicious manipulation. After that, we conduct ablation studies and give some further discussion. Section VI gives some conclusive results.

{
\section{Preliminaries}

\subsection{{Standard Adversarial} Perturbation}

\citet{szegedy2013intriguing,goodfellow2014explaining} show that deep neural networks (DNNs) obtained by {standard training} are vulnerable to adversarial perturbations, which means that adding imperceptible adversarial perturbation into the image may dramatically damage the output of DNNs.
Let $\mathbb{D}$ represent the space of natural images. Let $G$ represent the facial manipulation system which translates the input image to a new facial image. Given a natural image $\boldsymbol{x} \in \mathbb{D}$, the manipulated image can be expressed as $G(\boldsymbol{x})$. The adversarial perturbation $\boldsymbol{\delta}$ aims to distort the output of the model $G$, so that the adversarial attack can be formulated as the following maximum optimization problem:
\begin{equation}
     \max\limits_{||\boldsymbol{\delta}||_{p} \leq \epsilon} \mathcal{L}\left(G(\boldsymbol{x} + \boldsymbol{\delta}), G(\boldsymbol{x}) \right) ,
    \label{Eq:min-max_generative}
\end{equation} 
where $\mathcal{L}(\cdot)$ is a loss function to measure the difference between $G(\boldsymbol{x} + \boldsymbol{\delta})$ and $ G(\boldsymbol{x})$. $\boldsymbol{\delta}$ is the adversarial perturbation. $||\cdot||_{p}$ represents the $p$-norm and $\epsilon$ is the maximum perturbation constraint. 

Some existing methods \cite{ruiz2020disrupting,huang2021cmuawatermark} propose to use the MSE loss as the loss function and use the projected gradient descent (PGD) \cite{madry2019kpgd} to approximately solve this maximum optimization problem: 
\begin{footnotesize}
 \begin{equation}
 \left\{
 \begin{array}{l}   
    \boldsymbol{x_n} = \boldsymbol{x_{n-1}}+\eta\cdot sign(\nabla_{\boldsymbol{x_{n-1}}}MSE \left(G(\boldsymbol{x_{n-1}} + \boldsymbol{\delta}), G(\boldsymbol{x_{n-1}}) \right),\\
    \boldsymbol{x_n} = clip(\boldsymbol{x_n},\boldsymbol{x_0}-\epsilon,\boldsymbol{x_0}+\epsilon),
 \end{array}
 \right.
 \label{advpert_untargeted_generative}
 \end{equation}
 \end{footnotesize}
 where $\boldsymbol{x_n}$ is the generated adversarial example after $n$ steps, and $\boldsymbol{x_0}$ is the input image. {However, if the perturbation fails to disrupt the facial manipulation system, we need provenance tracing to distinguish whether the image is manipulated, which the standard adversarial perturbation can not provide.}
 Moreover, the initiative defense using standard adversarial perturbation can not distinguish whether the image is protected by the adversarial perturbation.

 \subsection{Information-containing Adversarial Perturbation}
 
 To address the limitations of {standard adversarial} perturbation, we propose our information-containing adversarial perturbation (\textbf{IAP}). \textbf{IAP} has two properties: on the one hand, it is an adversarial perturbation, which can distort the output of the facial manipulation systems; on the other hand, it can be decoded as an identity message to trace the source of the image in the database and detect the fake images.
 
 Given a natural image $\boldsymbol{x} \in \mathbb{D}$ and its identity message $M \in {\{0,1\}}^L$ of length $L$, we consider the task of perturbing the image as a transformation from the domain of natural images to the domain of adversarial images, seeking a mapping $E(\boldsymbol{x}, M) \rightarrow \boldsymbol{x}_{adv}$. The adversarial image $\boldsymbol{x}_{adv}$ needs to be perceptually similar to the input facial image $\boldsymbol{x}$ yet be able to distort the output of the facial manipulation system $G$, which means that $G(\boldsymbol{x}_{adv})$ should be very different from $G(\boldsymbol{x})$ while $d(\boldsymbol{x}_{adv}, \boldsymbol{x})$ is small ($d(\cdot, \cdot)$ is a distance metric). The mapping function $E(\boldsymbol{x}, M)$, which maps a natural image and its identity message to an adversarial image, can be parameterized using an end-to-end trainable encoder $E_{\theta_E}(\boldsymbol{x}, M)$. We train the encoder with the following optimization objective to ensure the effectiveness of the attack:
 \begin{equation}
 \begin{aligned}
 &\max\limits_{\theta_E} \mathbb{E}_{\boldsymbol{x}\in \mathbb{D}}[\mathcal{L}(G(E_{\theta_E}(\boldsymbol{x},M)), G(\boldsymbol{x}))] \\
 & \, s.t. \quad d(E_{\theta_E}(\boldsymbol{x}, M), \boldsymbol{x}) \leq \epsilon,
  \end{aligned}
 \end{equation}
 where $\mathcal{L}(\cdot)$ is a loss function to measure the difference between two images and the distance metric constrains the difference between the generated image $\boldsymbol{x}_{adv}$ and the original image $\boldsymbol{x}$.

 As for extracting the message contained in the perturbation, we train a decoder $D_{\theta_D}$ with the following optimization objective:
  \begin{equation}
 \min\limits_{\theta_D} \mathbb{E}_{\boldsymbol{x}\in \mathbb{D}}[{MSE}(D_{\theta_D}(\boldsymbol{x}_{adv}), M)],
 \end{equation}
 to approximate the mapping $D_{\theta_D}(\boldsymbol{x}_{adv}) \rightarrow M$. (See the Method Section for details.)

\subsection{Protecting Images Spread in Social Media Platforms with IAP}

Users can upload their photos to social media platforms (like Facebook and Twitter), sharing their lives with others. Unfortunately, evildoers can download these photos and apply off-the-shelf facial manipulation technology to produce a fake version of the user’s photos. As shown in Fig.\ref{img:intro}, the evildoer can successfully transform the photo of a young girl into an old woman. And then, the evildoer can upload this fake photo to the social media platform, violating the photo owner's privacy and portraiture right. That is to say, the unprotected social media platform is quite vulnerable in the scenario of identifying and preventing the spread of fake images.

Our \textbf{IAP} can establish a mechanism to protect facial images spread on the social media platform.
We construct a database with the facial image indexed by a unique message and refer to this message as the identity message of the image.

First, \textbf{IAP} can distinguish whether the image is protected by our method because the decoder can output an invalid message for the unprotected image. When the user uploads a facial image to the social media platform, the decoder of our method is invoked to check whether this facial image has been protected by \textbf{IAP}. {For an unprotected facial image (new image), \textbf{IAP} performs the following steps to protect this image: 1) generate a unique message as the index for this image and add this image into the database; 2) inject adversarial perturbation containing this unique message into the new facial image with the encoder.

Second, when the evildoer uploads the manipulated facial image to our social media platform, IAP performs the following steps to prevent the spread of this fake image: 1) recover the identity message in this image with the decoder; 2) trace the source of this image in the database using the recovered message as an index; 3) compare the uploaded image with the source image in the database and regard the uploaded image that is different from the source image as a fake image; 4) block the spread of the fake facial image. This idea is motivated by the barcode and QR code used to protect and track the source of the food.}

Third, \textbf{IAP} is a kind of adversarial perturbation that can distort the output of the facial manipulation systems, which makes the evildoer can hardly generate a realistic fake image. As shown in Fig.\ref{img:intro}, when the evildoer tries to transform the protected photo of a young girl into an old woman, the output of the facial manipulation system is distorted.

{In this way, our proposed \textbf{IAP} provides two-tier protection for the facial image, including initiatively attacking the facial manipulation systems to prevent malicious manipulation and passively tracing the source of the image to identify whether the image is fake. }

}

\section{Related Work}
This section briefly reviews the literature on deep face forgery and the existing defenses against facial manipulation.

 \subsection{Deep Face Forgery}
 Generative models (Auto-encoder \cite{kingma2013auto,razavi2019generating} and GANs \cite{goodfellow2014generative,richardson2021encoding}) have achieved tremendous progress in image synthesis, which are widely employed in creating modern DeepFakes. In the whole image synthesis scenario, generative models are used to synthesize new facial images that do not exist in the world. LAPGAN \cite{denton2015deep} and PGGAN \cite{karras2018progressive} can be used to generate high-resolution faces. StyleGAN \cite{karras2019style}, which utilizes the idea of style transfer \cite{luan2017deep}, has the ability to generate fake faces with specific attributes. In the partial synthesis scenario, generative models are used to automatically manipulate the face attributes like hair, gender, and expression for a given image. StarGAN \cite{choi2018stargan} takes the domain information and images together as input for training, and adds a mask vector to the domain label, so as to achieve the cross-domain style transfer. Fixed-point GAN \cite{siddiquee2019learning} proposed a new training method to alleviate artifacts in fake images: supervising the conversion of the same domain and regularizing the conversion of cross-domain. On the basis of CycleGAN \cite{zhu2017unpaired}, AttentionGAN \cite{tang2021attentiongan} generates images and attention maps simultaneously, and then fuses attention maps, generated images, and original images to get the final generated images. HiSD \cite{li2021imageHiSD} is a state-of-the-art image-to-image translation method that achieves multi-label scalability and controllable diversity through unsupervised disentanglement. There is no doubt that it would be offensive if these generative models spoofed our portrait photos.
 
 \subsection{Defense against facial manipulation}
 \textbf{Passive defense methods.}
 Detecting whether a facial image has been manipulated by generative models is a simple and straightforward idea. Some detection-based defense methods leverage the artifacts generated by generative models as clues and design different detectors with DNN-based classifiers.  Face X-ray \cite{li2020face} can detect fake images by showing the blending boundary for the fake image and the pure boundary for the real image. \citet{dang2020detection} focus on the tampered area in the facial images by adding an attention mechanism to the backbone network. \citet{li2020fighting} design a two-branch framework based on picture patches to improve detection ability through multi-task learning. The first branch learns the microscopic features of face blocks, and the other branch learns the differences between face and background areas. \citet{Multi_Scale_Texture} show that multi-scale texture difference can be utilized to improve face forgery detection.
 \citet{liu2021spatial} rethink the fake forgery detection from a frequency perspective and propose to combine spatial image and phase spectrum to capture the up-sampling artifacts of face forgery. \citet{wang2021representative} propose an attention-based data augmentation framework to guide the detector to refine and enlarge its attention, encouraging the detector to mine more profoundly into the regions ignored before for more representative forgeries.
 
 The physiological features can also be used to detect fake images.
 \citet{li2018ictu} observed the absence of eye blinks in synthetic faces, which could serve as a clue to detect fake videos. \citet{YangMWL21lip} propose a representative lip feature extraction and classification network to detect the fake image through the dynamic lip movement.
 \citet{fernandes2019predicting} propose to utilize bio-signals to distinguish fake videos. They extracted the heart rate by three methods: the color change of facial skin caused by blood flow, the average optical density of the forehead, and the change of the Euler image. They detect the fake video by the difference in heart rate distribution. 
 
 Digital watermarking is one of the important methods for digital multimedia copyright protection \cite{jiansheng2009digital,podilchuk2001digital,yavuz2007improved}. End-to-end watermark technology has shown advantages with the development of deep learning. HiDDeN \cite{zhu2018hidden} {proposes} the first end-to-end framework via jointly training encoders and decoders. For protecting facial privacy, \citet{wang2021faketagger} propose to embed the watermarking in the facial images to trace the source of the image.
 
 Passive defense can detect the fake image or trace the source of the image after malicious manipulation, but it cannot prevent malicious manipulation.
 
 \textbf{Initiative defense methods.}
 The adversarial perturbation always aims at increasing the classification loss \cite{goodfellow2014explaining,madry2019kpgd,wang2020unsupervised,zhong2020towards,yu2021salience} and changing the image to a different class. Some works introduce the adversarial into attacking generative models to protect facial images against manipulation.
 \citet{ruiz2020disrupting} propose a gradient-based method to attack facial manipulation systems with FGSM \cite{goodfellow2014explaining} and I-FGSM \cite{kurakin2016adversarial}. \citet{huang2021initiative} introduce a generator to improve the efficiency of generating adversarial perturbation. \citet{huang2021cmuawatermark} propose to attack multiple facial manipulation systems with a two-level perturbation fusion strategy to generate cross-model attacks. Injecting adversarial perturbation into the image can successfully prevent facial manipulation. However, these methods only restrict the perturbation within a $\epsilon$-ball which leads to low visual quality. Moreover, the standard adversarial perturbation seems like noise, which lacks useful information for provenance tracking.

 Our \textbf{I}nformation-containing \textbf{A}dversarial \textbf{P}erturbation (\textbf{IAP}) takes the essence from both the passive defense and the initiative defense and overcomes their disadvantages. On the one hand, \textbf{IAP} can initiatively disrupt facial manipulation systems. On the other hand, the information in the perturbation can serve for provenance tracking.

\section{Method}

\subsection{Architecture overview}
\begin{figure*}[t]
\centering
\includegraphics[width=1\textwidth]{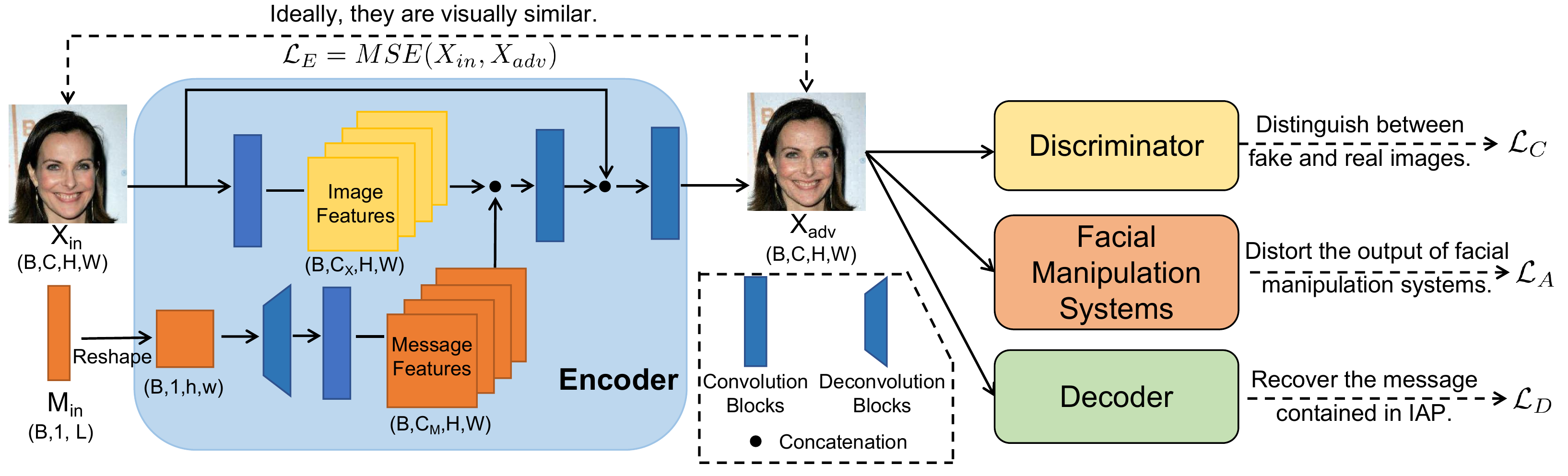}
\caption{Architecture overview. The encoder maps the original image $X_{in}$ and its corresponding message $M_{in}$ to an adversarial example $X_{adv}$. The discriminator is used to distinguish the original image and the adversarial image. The module "Facial Manipulation Systems" is used to ensure the effectiveness of adversarial examples. The decoder is used to extract the message contained in adversarial examples. The dotted arrows point to different components of losses. ``B" represents the batch size, ``C" represents the channels, ``H" represents the height of the image and ``W" represents the width of the image.}
\label{img:pipeline}
\end{figure*}

Considering that deep neural networks (DNNs) {have} made significant progress in various computer vision fields, we employ a DNN-based encoder and decoder to construct \textbf{i}nformation-containing \textbf{a}dversarial \textbf{p}erturbation (\textbf{IAP}). 
As shown in Fig.~\ref{img:pipeline}, the model architecture of our \textbf{IAP} contains four components. In the following, we describe the functionality of each component.

 \begin{itemize}
\item[$\bullet$] The encoder $E_{\theta_E}$ with parameters $\theta_E$ maps the original image and a corresponding message to an adversarial example. It receives the RGB input image $X_{in}$ in the shape of $3 * H * W$ and the binary message $M_{in} \in {\{0,1\}}^L$ of length L as input, and then generate the adversarial image $X_{adv}$ in the shape of $3 * H * W$.
 \item[$\bullet$] The discriminator $C_{\theta_C}$ with parameters $\theta_C$ works as an adversary to the encoder to guarantee the visual quality of the generated adversarial examples. It receives the image $X_{in}$ or $X_{adv}$ as input and predicts whether a given image is an adversarial example.
 \item[$\bullet$] The Facial manipulation systems $G$ are target models that can conduct face editing on a given image $X_{in}$.
  We aim to disrupt the generative capability of these systems with $X_{adv}$ by adding adversarial perturbations on $X_{in}$.
 \item[$\bullet$] The decoder $D_{\theta_D}$ with parameters $\theta_D$ extracts the message $M_{out}$ contained in the adversarial image. 
 \end{itemize}

\subsection{Encoder}
Our encoder aims to map the input image and its corresponding message to an adversarial example that is visually similar to the input image.
To begin with, the message $M$ of length $L$ is reshaped to ${\{0,1\}}^1*h*w$ where $L=1*h*w$. We use several deconvolution blocks with $stride=2$ to gradually amplify the message to the shape of $C_{M} * H * W$ where $C_M$ is the number of message feature channels. Then, we use convolution blocks to extract the features of the message and maintain the shape of the message features. 
Meanwhile, we extract the image features utilizing several convolution blocks. The image features are in the shape of $C_{X} * H * W$ where $C_X$ is the number of image feature channels.
Next, we concatenate the message features and the image features to a new tensor in the shape of $(C_{M}+C_{X}) * H * W$. 
Finally, this new tensor is concatenated with the input image $X_{in}$ and fed into a convolution block to generate the adversarial image $X_{adv}$.

One of the optimization goals of the encoder is to make the adversarial image and the input image visually similar. We minimize the mean square error between the adversarial image and the original image by updating the parameters $\theta_E$ as:

 \begin{equation}
 \begin{aligned}
    \mathop{\mathcal{L}_{E}}(X_{in},M_{in}) &\triangleq {MSE(X_{in},X_{adv})} \\
    &= {MSE(X_{in},E_{\theta_E}(X_{in},M_{in}))}.
 \label{LE1}
 \end{aligned}
 \end{equation} 

\subsection{Discriminator}
The discriminator $C_{\theta_C}$ consists of several convolution layers, a global average pooling layer, and a fully connected layer. The discriminator, which strives to distinguish input images from adversarial images, works as an adversary of the encoder. 
We update the parameters $\theta_C$ to maximize:

 \begin{equation}
 \begin{aligned}
     \mathop{\mathcal{L}_{C}}(X_{in},M_{in}) &\triangleq \log (1-C_{\theta_C}(E_{\theta_E}(X_{in},M_{in})))\\
    & + \log(C_{\theta_C}(X_{in})).
 \label{LC}
 \end{aligned}
 \end{equation} 
 
Conversely, the encoder expects the adversarial image to be close to the input image, preventing the discriminator from distinguishing adversarial images from original images. Thus we update the parameters $\theta_E$ to minimize $\mathcal{L}_{C}$.

\subsection{Attack Facial Manipulation Systems} \label{attack_facial}
In the previous two sections, we optimize the parameters of the encoder $E_{\theta_E}$ and discriminator $C_{\theta_C}$ so that the adversarial images produced by the encoder are close to the input images. In this section, we focus on how to make adversarial images effectively disrupt facial manipulation systems.

We denote the facial manipulation systems as $G$, the output on the original image as $G(X_{in})$, and the output on the adversarial image as $G(X_{adv})$.
The existing methods \cite{ruiz2020disrupting,huang2021initiative,huang2021cmuawatermark} always utilize Mean Square Error (MSE) to measure the differences between $G(X_{adv})$ and $G(X_{in})$:

 \begin{equation}
 \begin{aligned}
    \mathop{\mathcal{L}_{MSE}}(X_{in},X_{adv}) &\triangleq {MSE(G(X_{in}),G(X_{adv}))}.
 \label{LMSE}
 \end{aligned}
 \end{equation}
 These existing methods maximize the $\mathcal{L}_{MSE}$ in order to disrupt the performance of the facial manipulation systems. 
 
\begin{figure}[htbp!]
\centering
\includegraphics[width=0.495\textwidth]{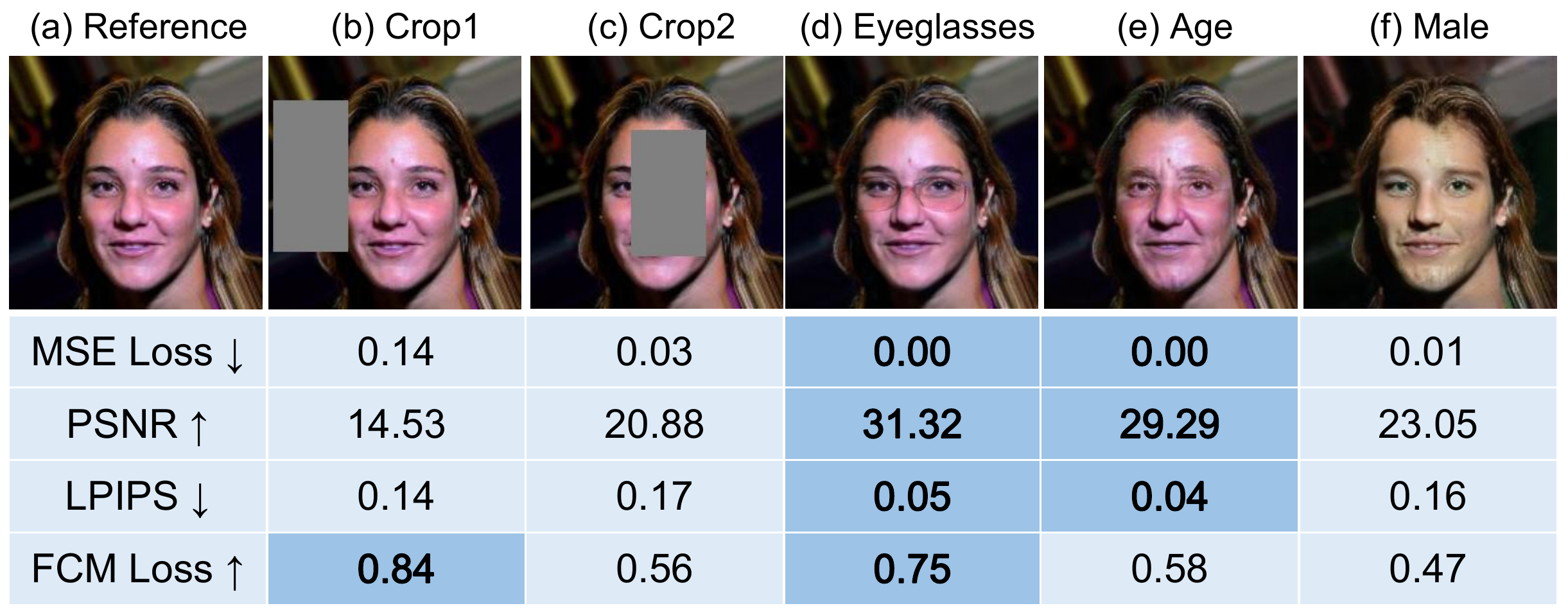}
\caption{Measure the similarity between different images and the reference image with MSE and our \textbf{f}eature-level \textbf{c}orrelation \textbf{m}easurement (FCM). We also show the results of another two perceptual metrics PSNR and LPIPS \cite{zhang2018unreasonableLPIPS}. The upward arrow means a larger metric indicates the closer similarity of the images and vice versa. We use dark blue to mark the two examples that are considered to be closest to the reference image under each metric. We round the results to two decimal places.}
\label{img:attackloss}
\end{figure}

 Here we first analyze the defects of this loss function in face protection and then propose our feature-level correlation measurement.
 The MSE loss is a pixel-level measurement for comparing the similarity between two images but lacks the comparison of the features in images. The MSE loss function can hardly capture changes in facial features and may be sensitive to changes other than the human face.
 As shown in Fig.\ref{img:attackloss}, the MSE loss indicate that the Fig.\ref{img:attackloss}(c) is closer to the reference image than the Fig.\ref{img:attackloss}(b), which does not meet our expectations. Although the MSE value for Fig.\ref{img:attackloss}(b) is relatively large, the features of the face are still clear, while the MSE value for Fig.\ref{img:attackloss}(c) is relatively small, but some facial features are successfully erased and can protect the privacy of this image.
 It is not very rigorous to use MSE as a metric to measure the performance of facial manipulation systems. In the ablation study, we show that using MSE loss to attack facial manipulation systems can not always generate effective information-containing adversarial examples.

From the perspective of protecting the reputation and privacy of facial images, we suggest comparing the features of the two images.
The features of a facial image include attributes such as gender, skin color, hair color, age, and so on. Comparing two images according to the attributes of the facial image is more in line with the visual perception of humans. 
If the feature similarity between the manipulated adversarial image and the manipulated original image is low, we think that the adversarial perturbation successfully disrupts the performance of the facial manipulated systems. We propose our \textbf{F}eature-level \textbf{C}orrelation \textbf{M}easurement (FCM). To start with, we train a classification model (ResNet-50 \cite{he2016deep}) for face attribute recognition on CelebA \cite{liu2015faceattributes} and use the model's penultimate layer as the feature extractor $F$. Then we use the cosine similarity to measure the correlation between the features. Fig.\ref{img:attackloss} shows that the feature-level correlation measurement gives more reasonable predictions on the perceptual similarity of facial images and is suitable for facial image protection. 
In order to disrupt the performance of the facial manipulation systems $G$ with the adversarial perturbation, we update the parameters $\theta_E$ to minimize the similarity between the feature of $G(X_{in})$ and $G(X_{adv})$:

 \begin{equation}
 \begin{aligned}
    \mathop{\mathcal{L}_{A}}(X_{in},M_{in}) &\triangleq Cos(F(G(X_{in})),F(G(X_{adv})))\\
    &=Cos(F(G(X_{in})),F(G(E_{\theta_E}(X_{in},M_{in})))).
 \label{LA}
 \end{aligned}
 \end{equation}

 \subsection{Decoder} \label{decoder}
 The decoder module $D_{\theta_D}$ turns the adversarial image into the tensor in the shape of $1*h*w$ through several convolution blocks, and then reshapes this tensor to the output message $M_{out} \in {\{0,1\}}^L$. The optimization objective of the decoder is to minimize the mean square error of the output message $M_{out}$ and input message $M_{in}$ by updating parameters $\theta_D$ as:
 \begin{equation}
  \begin{aligned}
    \mathop{\mathcal{L}_{M}}(X_{in},M_{in}) &\triangleq {MSE(M_{in},M_{out})} \\
    &={MSE(M_{in},D_{\theta_D}(X_{adv}))}.
 \label{LM}
 \end{aligned}
 \end{equation}

 Our \textbf{IAP} provides two-tier protection for faces, including attacking malicious manipulation and tracking the source of the adversarial image.
 For this reason, the message contained in adversarial examples should be robust to malicious manipulation, and the decoder needs to correctly extract the message from both the adversarial examples and the manipulated adversarial examples. However, we find that malicious manipulation may destroy the information contained in adversarial examples, making it challenging to extract the message. 
 
 \begin{figure}[htp]
\centering
\includegraphics[width=0.495\textwidth]{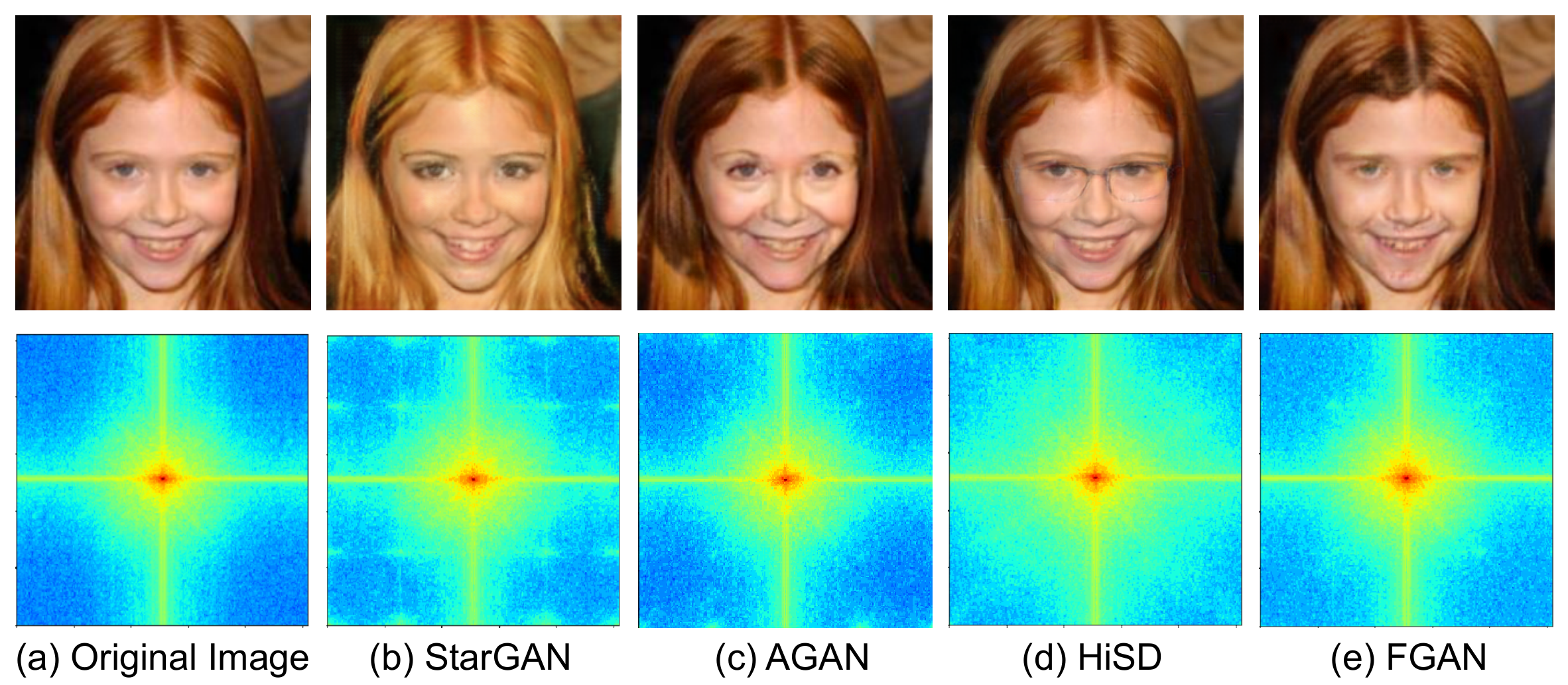}
\caption{The spectrogram of different images. Different colors indicate the intensity of the corresponding frequency. (a) The original image and its spectrogram. (b-e) The manipulated images through StarGAN (blond hair), HiSD ({wearing} glasses), AttentionGAN (old), and Fixed-point GAN (gender) and their spectrograms.}
\label{img:fre}
\end{figure}

 The way to improve the robustness of the hidden information in the common watermark-related literature \cite{zhu2018hidden,jia2021mbrs} is to add corresponding noise layers in the training process. For example, adding a blur noise layer in the training process contributes to resisting blur corruption, and adding a JPEG compression noise layer contributes to resisting JPEG compression. 
 
From the perspective of the influence of malicious manipulation on the frequency domain of images, we propose a novel spectral diffusion module to improve hidden information's robustness. We do not work to resist a specific corruption but take the commonality of these corruption methods into consideration.
 In Fig. \ref{img:fre}, we illustrate the spectrogram of the images (via Fourier transform) to represent the energy at different frequencies in the image. The low-frequency components are near the center of the spectrogram, and the high-frequency components are far away from the center. 
 Obviously, facial manipulation systems affect the frequency distribution of the image to varying degrees, which means that the information in the corresponding frequency channels may be disturbed. 
 HiSD and StarGAN have a greater impact on the spectrogram than AttentionGAN and Fixed-point GAN, and in experiments, HiSD and StarGAN greatly reduce the accuracy of message extraction, while AttentionGAN and Fixed-point GAN slightly reduce the accuracy of message extraction.
 Similarly, image corruptions such as compression and blurring will also damage the frequency channels to some extent. This observation motivates us to propose a \textbf{s}pectral \textbf{d}iffusion module (SD) that can be used in the end-to-end training scheme. By encouraging the diffusion of messages on different frequency channels, we reduce the impact of corruption on the frequency channels on message extraction. 
 
 In particular, the spectral diffusion module first performs color space transformation and discrete cosine transformation (DCT) on patches in the shape of $8*8$ of the image $X_{adv}$, which coincides with the JPEG compression standard \cite{pennebaker1992jpeg}. 
 We use the even symmetric DCT with 64 basis functions:
 
 \begin{small} 
 \begin{equation}
 \begin{aligned}
    F(u,v)=\frac{1}{4}\beta_{uv}[\sum\limits_{x=0}^7 \sum\limits_{y=0}^7 f(x,y)cos{\frac{(2x+1)u\pi}{16}}cos{\frac{(2y+1)v\pi}{16}}],
 \label{eq:DCT}
 \end{aligned}
 \end{equation}
 \end{small} 
 
 {\noindent where $\beta_{uv}=\frac{1}{2}$ when $u=v=0$, and $\beta_{uv}=1$ in other cases.
 Here, we get a DCT map for the patches of $X_{adv}$. And then the DCT map is divided by a random quantization table with values from 1 to 100. After that, the spectral diffusion module scales down the value that is less than 1 in the DCT map using a differentiable cubic power function, which aims to discard some information in these frequency channels. At last, the inverse DCT and color space transformation are applied to the DCT map to reconstruct the images. }
 
 {In the training process, the original image $X_{in}$ is mapped to an adversarial example $X_{adv}$ by the encoder, and then the information of some frequency channels is discarded by the \textbf{s}pectral \textbf{d}iffusion module (SD) and obtain a new adversarial example. We train the decoder to recover the message in this new adversarial example. Thus the optimization objective of the decoder turns to:}
 \begin{equation}
  \begin{aligned}
    \mathop{\mathcal{L}_{D}}(X_{in},M_{in}) &\triangleq  MSE(M_{in},D_{\theta_D}(SD(E_{\theta_E}(X_{in},M_{in})))).
 \label{LD}
 \end{aligned}
 \end{equation}
 
In this way, the encoded information is encouraged to diffuse into different channels, and destroying some frequency channels will not affect the message extraction performance of the decoder. {It's worth noting that this module is used to improve the robustness of the message extraction in training the encoder and decoder and is not used in the inference process.} Our \textbf{IAP} is robust to facial manipulation. Moreover, \textbf{IAP} shows satisfying robustness against JPEG compression even surpassing the existing methods which are specially designed against JPEG \cite{zhu2018hidden,liu2019novel, jia2021mbrs} (See results in Sec.\ref{sec:extraction}).
 
 \textbf{Difference between the robustness improvement method in watermark literature \cite{liu2019novel,huang2021cmuawatermark} and our method}. The process of our spectral diffusion module includes color space transformation, discrete cosine transformation and quantization operations, which is similar to JPEG compression.
 For that JPEG compression is an essential and commonly used image processing operation, there is much research on improving the watermark's robustness against the JPEG \cite{zhu2018hidden,xing2021invertible,huang2021cmuawatermark}. \textbf{But why {can these methods} not resist facial manipulation systems?} 
 The existing methods intuitively train with the different noise layers to guarantee robustness to the image corruption. They simulate the process of the JPEG and deal with the non-differential problem in JPEG compression from different perspectives. In view of the insensitivity of humans to high-frequency regions, JPEG filters out the high-frequency parts by dividing the image by a specific quantization table. The data in this table are based on the sensitivity of humans to different frequencies. As a comparison, in order to improve the robustness against the damage not only in some high-frequency channels but also in other frequency channels, we use the random quantization table rather than the specific quantization table in the spectral diffusion module. In this way, we randomly discard the information contained in different frequency channels in the training process, while the existing methods just consider discarding the high-frequency channels. We show that our spectral diffusion module can greatly improve the robustness against facial manipulation and common image corruptions in experiments.

 The optimization problem of our method can be formulated as a min-max optimization (like GANs). Let $\mathbb{D}$ represent the training dataset and $\mathbb{M}$ represent the distribution of the messages. We sample a facial image $\boldsymbol{x} \in \mathbb{D}$ and a message $M \in \mathbb{M}$. We first perform stochastic gradient ascent on the parameters $\theta_C$ to maximum loss $\mathcal{L}_C$ to distinguish input images from adversarial images. Then we perform stochastic gradient descent on the parameters $\theta_E$ and $\theta_D$ to minimize the loss $\mathcal{L}(\boldsymbol{x},M)=\lambda_{E} \mathcal{L}_{E}(\boldsymbol{x},M)+ \lambda_{C} \mathcal{L}_{C}(\boldsymbol{x},M)+\lambda_A\mathcal{L}_{A}(\boldsymbol{x},M)+\lambda_D\mathcal{L}_{D}(\boldsymbol{x},M)$ to make the discriminator can hardly distinguish input images from adversarial images, the encoder can effectively generate the IAP adversarial image and the decoder can correctly recover the message contained in the adversarial image. The formulation of this optimization problem can be expressed as:
 
 \begin{small}
 \begin{equation}
 \begin{aligned}
 \min\limits_{\theta_E,\theta_D} \max\limits_{\theta_C} \mathbb{E}_{\boldsymbol{x}\in \mathbb{D},M\in\mathbb{M}}[&\lambda_{E} \mathcal{L}_{E}(\boldsymbol{x},M)+ \lambda_{C} \mathcal{L}_{C}(\boldsymbol{x},M)\\
 &+\lambda_A\mathcal{L}_{A}(\boldsymbol{x},M)+\lambda_D\mathcal{L}_{D}(\boldsymbol{x},M)]. 
 \end{aligned}
 \end{equation}
 \end{small}
 Our algorithm is outlined in Algorithm \ref{algorithm:algorithm}. {In the training process, we sample messages randomly at each bit to ensure that our method can extract different messages correctly, following the setting in the robust digital watermark \cite{zhu2018hidden,jia2021mbrs}. When using IAP to protect images in social media platforms, the message is required to be unique for each image, as this message is used as an index in the database for provenance tracking.}
 
 \begin{algorithm}[htbp]
    \caption{{IAP Training: Given the training dataset $\mathbb{D}$, epochs $T$, batch size $B$, learning rate $\eta$. We optimize the parameters $\theta_E$ for the encoder $E_{\theta_E}$, $\theta_D$ for the decoder $D_{\theta_D}$ and $\theta_C$ for the discriminator $C_{\theta_C}$. $\lambda$ represents the hyperparameter to adjust the weight of the loss.}}
    \begin{algorithmic}

    \For {$i = 1,2...,T$}
        
        \State Sampling minibatch of images $\{$$\boldsymbol{x}^{(1)}$, ..., $\boldsymbol{x}^{(B)}$$\}$ from the training dataset $\mathbb{D}$. Sampling the message randomly at each bit.
        \State $\blacktriangleright$ Updating the discriminator $C_{\theta_C}$ by ascending the stochastic gradient:
        \State $\theta_C=\theta_C+\eta \cdot (\nabla_{\theta_C}\mathcal{L}_C)$ 
        \State $\blacktriangleright$ Updating the encoder $E_{\theta_E}$ and decoder $D_{\theta_D}$ jointly:
        
        \State $\left[\theta_E ;\theta_D\right] =\left[ \theta_E ; \theta_D\right]- \eta \cdot\nabla_{\theta_E, \theta_D}(\lambda_{E} \mathcal{L}_{E}+ \lambda_{C}\mathcal{L}_{C}+\lambda_A\mathcal{L}_{A}+\lambda_D\mathcal{L}_{D})$
	\EndFor

  \State
    \end{algorithmic}
    \label{algorithm:algorithm}
\end{algorithm}

\section{Experiment}

\subsection{Implementation}
\textbf{Details.} In our experiments, we use the CelebA \cite{liu2015faceattributes} as the main dataset and train the \textbf{IAP} model on the training set of CelebA. We evaluate our method on the testing set of the CelebA and the \textbf{LFW} \cite{LFWTech} dataset to ensure the generalization of the trained \textbf{IAP} model. The facial manipulation systems chosen in our experiments are StarGAN \cite{choi2018stargan}, AttentionGAN (noted as AGAN in table) \cite{tang2021attentiongan}, Fixed-point GAN (noted as FGAN in table) \cite{siddiquee2019learning}, and HiSD \cite{li2021imageHiSD}.
StarGAN, AttentionGAN, and Fixed-point GAN are all trained on the CelebA dataset for five attributes: black hair, blond hair, brown hair, gender, and age. The latest facial manipulation system HiSD that can add a pair of glasses to the target face is also considered in our experiments. The framework of \textbf{IAP} is implemented by PyTorch \cite{pytorch} and runs on RTX-2080TI. For a fair comparison, the models in this paper are trained with the same image size (Height=256, Width=256). The length of the message contained in the adversarial example is $L=64$ and the message is sampled randomly at each bit in training. For the weight factors of the loss function, we choose $\lambda_{E}=1$, $\lambda_{C}=0.0001$, $\lambda_D=10$ and $\lambda_A=0.0025$.
As for the stochastic optimization method, we use Adam \cite{kingma2014adam} with a learning rate of $10^{-3}$. The batch size is set to 4, and the maximum training epoch is 100.

\textbf{Metrics.}
\textbf{IAP} provides two-tier protection for facial images, including initiative protection to disrupt the facial manipulation system and passive protection to trace the source of images after malicious manipulation.
Thus there are three main indicators to evaluate our model: image quality, initiative attack success rate, and message extraction accuracy.

We choose two traditional metrics PSNR \cite{almohammad2010stegoPSNR} and SSIM \cite{wang2004imageSSIM} and one deep network based metric LPIPS \cite{zhang2018unreasonableLPIPS} to measure the image quality. Higher LPIPS means a larger difference between the two images, while lower PSNR (SSIM) means a larger difference. The LPIPS metric is thought to correlate well with human perception of image quality. The adversarial example should be visually similar to the original image.

\begin{figure}[htbp!]
\centering
\includegraphics[width=0.49\textwidth]{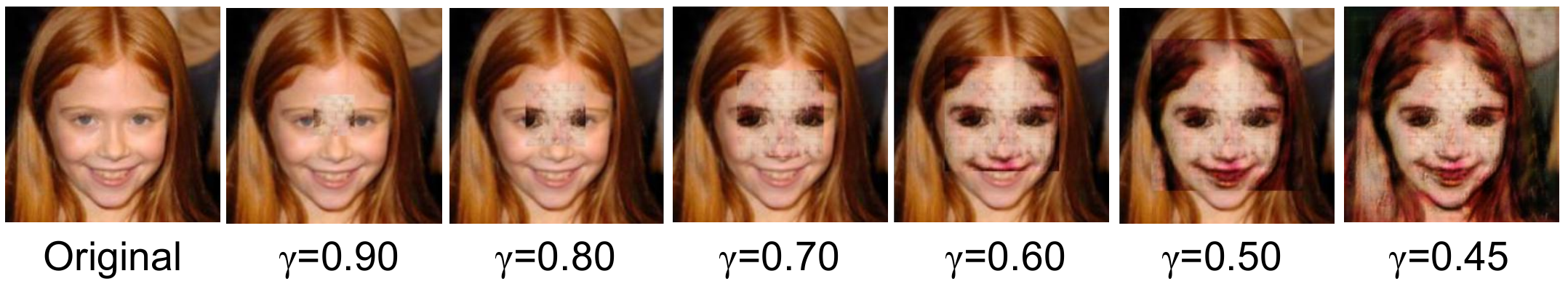}
\caption{The feature-level correlation measurement (FCM) between the original image and the corrupted images. We think that the image at $FCM=0.7$ is already distorted and different from the original image. We set the threshold $\gamma=0.7$ to evaluate the initiative attack success rate.}
\label{img:gamma}
\end{figure}

{
For initiative \textbf{a}ttack \textbf{s}uccess \textbf{r}ate (\textbf{ASR}), we consider measuring the difference between the output of the facial manipulation system on the adversarial image and the original image. If the output of the facial manipulation system on an adversarial image is perceptually different from that on the original image, we regard this adversarial example as a successful adversarial example.
\citet{zhang2018unreasonable} demonstrate the effectiveness of using the features of deep models to measure perceptual similarity, which motivates us to use the penultimate layer of a ResNet-50 \cite{he2016deep} (pre-trained for face attribute recognition) as a feature extractor and measure the perceptual similarity by comparing the cosine similarity between the features of different images.
This kind of \textbf{f}eature-level \textbf{c}orrelation \textbf{m}easurement (FCM) can be expressed as:
\begin{equation}
\label{sr}
\begin{aligned}
FCM(\boldsymbol{x}_{1},\boldsymbol{x}_{2})=Cos(F(\boldsymbol{x}_1),F(\boldsymbol{x}_2)),
\end{aligned}
\end{equation}
where $F$ represents the feature extractor.
We regard the facial image as successfully protected by the adversarial perturbation when:
\begin{equation}
\label{sr_}
\begin{aligned}
FCM(G(X_{in}),G(X_{adv}))\leq \gamma,
\end{aligned}
\end{equation}
where $\gamma$ is a threshold. We choose $\gamma=0.7$ in our experiments. (Visually, we illustrate some images with different FCM in Fig. \ref{img:gamma}.)}

As for message extraction accuracy, we introduce the Bit Error Rate (BER) as the measurement. 

\textbf{Baselines.} As a two-tier protection scheme, \textbf{IAP} should consider the baselines from two aspects for comparison. For the initiative protection, we choose three state-of-the-art methods I-FGSM \cite{ruiz2020disrupting}, PG \cite{huang2021initiative}, and CMUA \cite{huang2021cmuawatermark} for comparison. I-FGSM is a gradient-based method to attack the facial manipulation system. PG is a generator-based method to generate adversarial perturbation. CMUA is a universal perturbation that can be applied to many facial images. 

As for extracting the information contained in \textbf{IAP} to trace the source of the image, there are few related references in the field of face protection. For comparison, we choose JPEG-Mask \cite{zhu2018hidden}, TSR \cite{liu2019novel}, and MBRS \cite{jia2021mbrs}, which are deep learning based robust watermarking schemes, as baselines. JPEG-Mask proposes to simulate the JPEG process in the one-stage end-to-end training process. TSR proposes a novel two-stage separable framework that considers the real JPEG compression in the training process. MBRS proposes randomly utilizing real and simulated JPEG compression during training and is the SOTA robust watermarking method.

 \begin{table*}[htbp]
\centering
\caption{The quantitative experiments for initiative defense. The best results are in bold. We use SSIM, PSNR and LPIPS to measure the perceptual similarity between the adversarial examples and the original. The FCM in the table represents the feature-level correlation measurement, and the ASR represents the attack success rate for disrupting the facial manipulation systems.} 
\scalebox{0.96}{
\begin{tabular}{c|c|c|c|c|c|c|c|c}
\hline \hline
Dataset & \multicolumn{1}{c|}{Method} & \multicolumn{1}{c|}{SSIM} & \multicolumn{1}{c|}{PSNR} & \multicolumn{1}{c|}{LPIPS} & \multicolumn{1}{c|}{HiSD(ASR) \cite{li2021imageHiSD}} & \multicolumn{1}{c|}{StarGAN(ASR) \cite{choi2018stargan}} & \multicolumn{1}{c|}{AGAN(ASR) \cite{tang2021attentiongan}} & \multicolumn{1}{c}{FGAN(ASR) \cite{siddiquee2019learning}} \\ \hline
\multirow{4}{*}{CelebA \cite{liu2015faceattributes}} & I-FGSM \cite{ruiz2020disrupting} & 0.7643 & 35.93 & 0.1135 & 1.0000 & 1.0000 & 0.9833 & 1.0000 \\
 & PG \cite{huang2021initiative}& 0.7360 & 35.77 & 0.1172 & 0.9529 & 0.9922 & 0.9606 & 0.9968 \\
 & CMUA \cite{huang2021cmuawatermark}& 0.7412 & 35.89 & 0.1058 & 0.9827 & 0.9968 & 0.9648 & 0.9881 \\ \cline{2-9} 
 & IAP (ours)& \textbf{0.8524} & \textbf{36.16} & \textbf{0.1053} & 0.9883 & \textbf{1.0000} & \textbf{0.9850} & 0.9984 \\
  & FCM-attack (ours)& \textbf{0.9702} & \textbf{45.92} & \textbf{0.0342} & \textbf{1.0000} & \textbf{1.0000} & \textbf{1.0000} & \textbf{1.0000} \\
 \hline
\multirow{4}{*}{LFW \cite{LFWTech}} & I-FGSM \cite{ruiz2020disrupting} & 0.7483 & 35.92 & 0.1274 & 1.0000 & 1.0000 & 0.9816 & 0.9950 \\
 & PG \cite{huang2021initiative}& 0.7154 & 35.76 & 0.1275 & 0.9400 & 0.9886 & 0.9633 & 0.9850 \\
 & CMUA \cite{huang2021cmuawatermark}& 0.717 & 35.85 & 0.1142 & 0.9950 & 1.0000 & 0.9650 & 0.9870 \\ \cline{2-9} 
 & IAP (ours)& \textbf{0.8482} & \textbf{37.17} & \textbf{0.1032} & 0.9900 & 1.0000 & \textbf{0.9833} & 0.9911 \\
  & FCM-attack (ours)& \textbf{0.9522} & \textbf{44.91} & \textbf{0.0353} & \textbf{1.0000} & \textbf{1.0000} & \textbf{1.0000} & \textbf{1.0000} \\\hline \hline
\end{tabular}\label{tab:attack_results}}
\end{table*}

\begin{figure}[htbp!]
\centering
\includegraphics[width=0.49\textwidth]{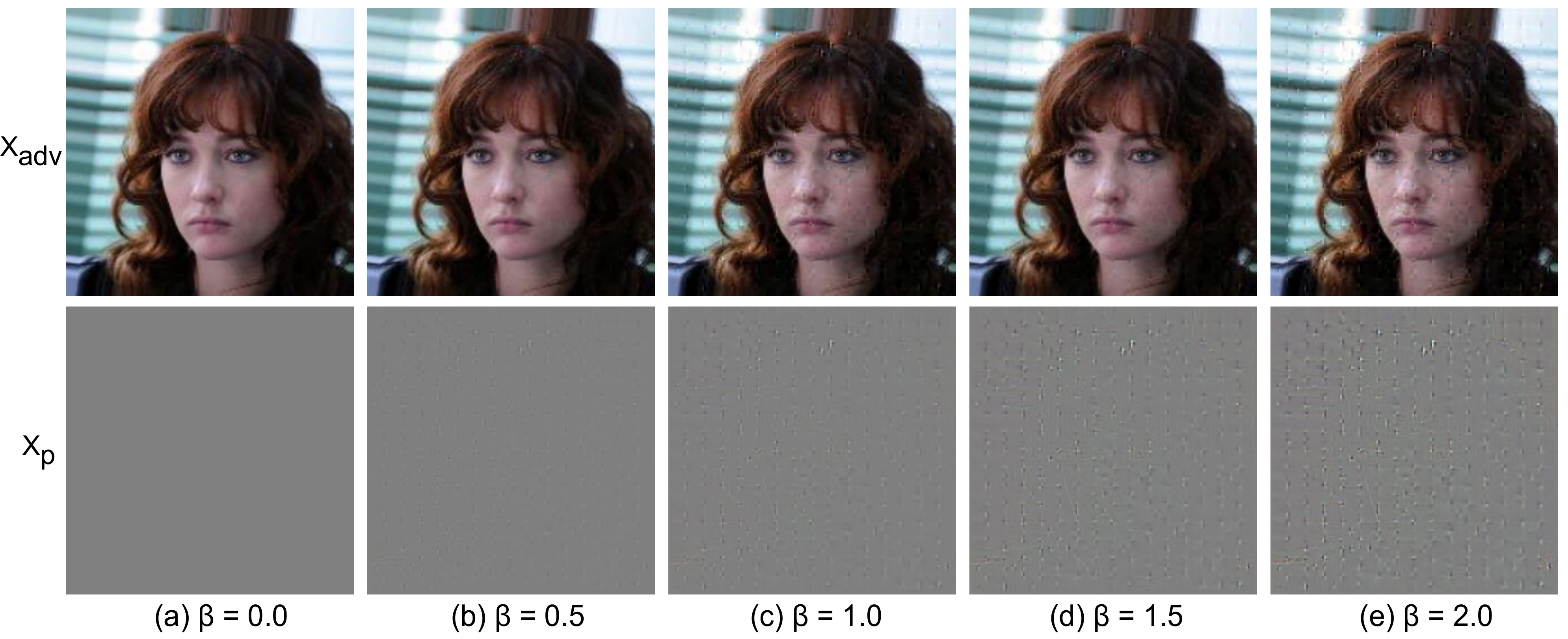}
\caption{We illustrate the adversarial examples and the adversarial perturbations with different $\beta$. Reducing the perturbation strength is helpful in improving the visual quality of the adversarial examples. }
\label{img:beta}
\end{figure}

  \textbf{Perturbation Strength.} We refer to the residual between the input image $X_{in}$ and the adversarial example $X_{adv}$ as the adversarial perturbation $X_{p}$:
 \begin{equation}
  \begin{aligned}
    X_p = X_{adv} - X_{in}.
 \label{pert}
 \end{aligned}
 \end{equation}
 In the inference stage, we introduce a factor $\beta$ to adjust the trade-off between the strength of adversarial perturbation and the visual quality of the adversarial example:
  \begin{equation}
  \begin{aligned}
    X_{badv} = X_{in} + \beta X_{p}.
 \label{beta}
 \end{aligned}
 \end{equation}
 In Fig.~\ref{img:beta}, we show the adversarial examples and their corresponding adversarial perturbation with different perturbation strengths $\beta$. Adversarial examples with different perturbation strengths can be obtained by adjusting the parameter $\beta$. For {fairness}, we compare different methods with the same image quality in the experiments by adjusting $\beta$.

\subsection{Initiative Protection}
\begin{figure}[htbp!]
\centering
\includegraphics[width=0.49\textwidth]{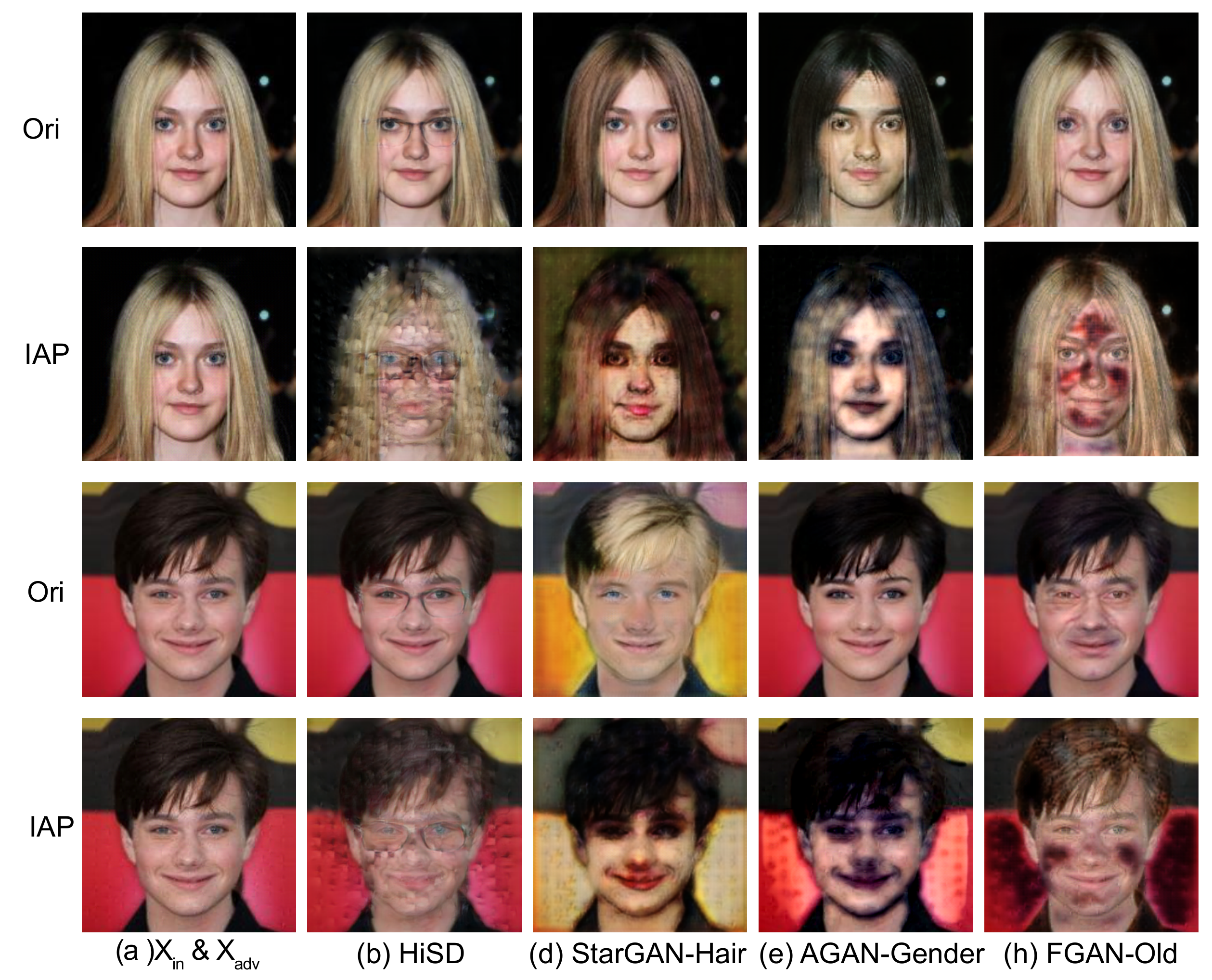}
\caption{The qualitative experiments for initiative defense. We illustrate the original images $X_{in}$ and the manipulated output on $X_{in}$ in the first and third lines. We also illustrate the \textbf{IAP} examples $X_{adv}$ and their manipulated counterparts in the second and the fourth lines.
The facial manipulation systems can successfully manipulate the original images, such as adding eyeglasses with HiSD, changing the hair color with StarGAN (SGAN), changing the gender with AttentionGAN (AGAN) and changing the age with Fixed-point GAN (FGAN). Our IAP can effectively disrupt different manipulation systems.}
\label{img:attack_result}
\end{figure}

We conduct qualitative and quantitative experiments to demonstrate the effectiveness of the proposed \textbf{IAP} in the initiative defense for facial images.
As shown in Fig.\ref{img:attack_result}, the facial manipulation systems can successfully manipulate the original image while \textbf{IAP} initiatively disrupts the performance of these systems. \textbf{IAP} generate one information-containing adversarial example for an image and use this adversarial example to attack different facial manipulation systems. Visually, \textbf{IAP} can protect the facial image from malicious manipulation because the output of the facial manipulation systems on the image protected by \textbf{IAP} lacks valid face features and can easily be identified as a fake image.

The quantitative experiments are reported in Tab.\ref{tab:attack_results}. 
In practice, I-FGSM \cite{ruiz2020disrupting} is an image-specific method that needs to generate adversarial examples iteratively for each image against one specific facial manipulation model. Thus, I-FGSM performs well for disrupting a specific model (as shown in Tab.\ref{tab:attack_results}), at the cost of time for generating adversarial examples. PG \cite{huang2021initiative} trains a generator against a specific model that can generate adversarial examples with a forward pass only. PG can obtain adversarial examples more efficiently, while the attack success rate is slightly lower than I-FSGM. CMUA \cite{huang2021cmuawatermark} {proposes} to generate one cross-model universal perturbation for attacking different facial manipulation systems with a perturbation fusion strategy. Our \textbf{IAP} is also a cross-model attack and can generate adversarial examples with a forward pass only. We only generate one adversarial example for an image, and use this adversarial example to disrupt different facial manipulation systems. The "FCM-attack" in the table means the method that just uses our FCM loss to generate adversarial examples against a specific manipulation system without containing any information in the examples, which follows the experimental setting of I-FGSM.

\textbf{IAP} can achieve a competitive attack success rate compared with the I-FGSM attack.
In attacking AttentionGAN on CelebA, the attack success rate of \textbf{IAP} is $2.02\%$ higher than CMUA, $2.44\%$ higher than PG. In attacking Fixed-point GAN on CelebA, the attack success rate of \textbf{IAP} is $1.03\%$ higher than CMUA and $0.16\%$ higher than PG. 
FCM-attack can achieve both the best perception quality and the best attack success rate, which is in line with the analyze in Sec.\ref{attack_facial} that FCM is more suitable in face protection scenarios than the MSE loss.
Furthermore, \textbf{IAP} can not only effectively protect the facial images in CelebA, but also generalize well on the LFW dataset.

\subsection{Passive Protection} \label{sec:extraction}

\textbf{IAP} provides two-tier protection for facial images. On the one hand, \textbf{IAP} disrupts the facial manipulation systems as shown in Fig.\ref{img:attack_result} and makes such systems can not generate vivid facial images on the information-contained adversarial examples. On the other hand, an essential difference between Information-containing Adversarial Perturbation (\textbf{IAP}) and {standard adversarial} perturbation is that \textbf{IAP} is a meaningful perturbation that can be used as a clue for provenance tracking rather than messy noise. 
In this section, we conduct experiments to explore the effectiveness of \textbf{IAP} in extracting the message contained in our adversarial examples against different manipulation systems, which can be served as a passive protection method. 

In Tab.\ref{tab:extaction_results}, we compare our \textbf{IAP} with three SOTA robust watermark method, including JPEG-Mask \cite{zhu2018hidden}, TSR \cite{liu2019novel} and MBRS \cite{jia2021mbrs}. The method 'Identity' represents standardly training the encoder and decoder without our spectral diffusion module (in Sec.\ref{decoder}) and other noise layers used in the compared methods. All the methods can $100\%$ correctly extract the message contained in adversarial examples. Manipulating images by employing HiSD and StarGAN will greatly reduce the accuracy of message extraction, while AttentionGAN and Fixed-point GAN have a slightly negative influence on message extraction. As shown in Fig.\ref{img:fre}, AttentionGAN has a lighter effect on the image spectrogram than HiSD. The average bit error rate of our \textbf{IAP} against HiSD on CelebA is $1.11\%$ which is $9.88\%$ lower than MBRS and $47.25\%$ lower than the 'Identity' training method. Moreover, \textbf{IAP} performs well when dealing with StarGAN manipulation. The average BER for \textbf{IAP} against StarGAN on CelebA is $10.95\%$ lower than MBRS and $46.24\%$ lower than the 'Identity' training method. \textbf{IAP} can also generalize on the LFW dataset and surpass the existing methods against different facial manipulation methods.

\begin{table}[htb]
\caption{The average BER for extracting messages from the adversarial examples and their manipulated counterparts. We choose HiSD, StarGAN, AGAN and FGAN for facial manipulation. 'Adv.' represents the BER on adversarial examples. The method 'Identity' represents standardly training the encoder and decoder without any robust improvement. PSNR is adjusted to about 39 by the perturbation strength factor $\beta$ for fair. The best results are in bold.} 
\centering
\scalebox{0.9}{
\begin{tabular}{c|c|c|c|c|c|c}
\hline \hline
\multicolumn{1}{c|}{Dataset} & Method & Adv. & HiSD & StarGAN & AGAN & FGAN \\ \hline
\multirow{5}{*}{CelebA} & Identity & 0.00 & 48.36 & 46.83 & 12.39 & 4.75 \\ 
 & JPEG-MASK & 0.00 & 24.70 & 26.61 & 0.00 & 0.00 \\
 & TRS & 0.00 & 2.21 & 16.92 & 0.00 & 0.00 \\
 & MBRS & 0.00 & 10.99 & 11.54 & 0.00 & 0.00 \\
 & \textbf{IAP} & \textbf{0.00} & \textbf{1.11} & \textbf{0.59} & \textbf{0.00} & \textbf{0.00} \\ \hline
\multirow{5}{*}{LFW} & Identity & 0.00 & 48.65 & 45.94 & 10.93 & 4.68 \\
 & JPEG-MASK & 0.00 & 21.24 & 24.17 & 0.00 & 0.00 \\
 & TRS & 0.00 & 1.99 & 12.71 & 0.00 & 0.00 \\
 & MBRS & 0.00 & 2.30 & 4.87 & 0.00 & 0.00 \\
 & \textbf{IAP} & \textbf{0.00} & \textbf{0.94} & \textbf{0.60} & \textbf{0.00} & \textbf{0.00} \\ \hline \hline
\end{tabular}
\label{tab:extaction_results}}
\end{table}

\begin{figure}[htbp]
\centering
\includegraphics[width=0.49\textwidth]{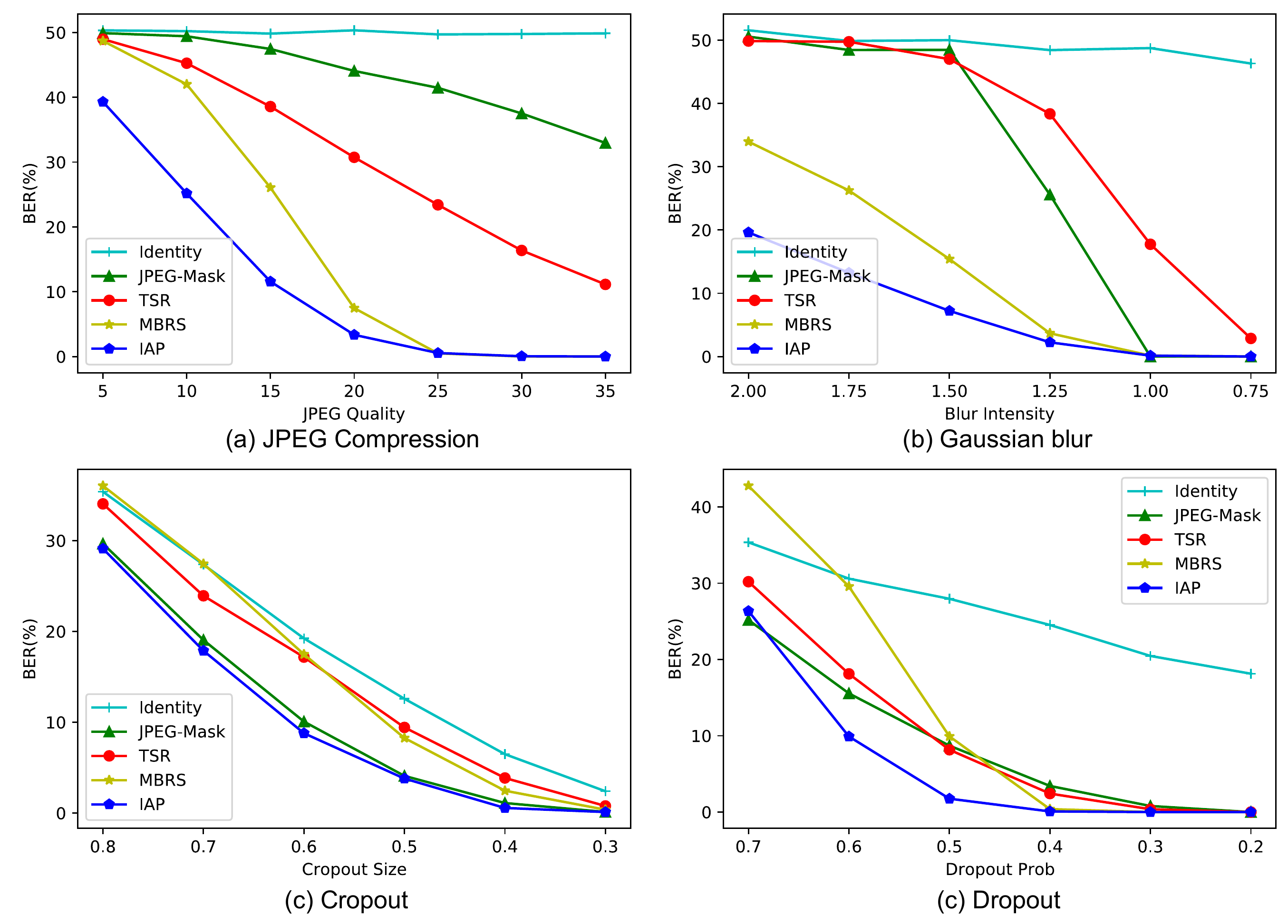}
\caption{The line graph compares the robustness of different methods against image corruptions with different strengths, and \textbf{IAP} surpasses the existing robust watermarking algorithms. PSNR is adjusted to about 39 by the perturbation strength factor $\beta$ for fair.}
\label{img:jpeg}
\end{figure}

In addition, we observe that our \textbf{IAP} is not only robust to different facial manipulation but also robust to other image corruptions like JPEG compression, blur, cropout $etc.$ In Fig.\ref{img:jpeg}, we illustrate the BER of different methods against image corruptions with different strengths on the CelebA. JPEG quality 10 (JPEG(10)) means the amount of compression is 90$\%$. The result in Fig.\ref{img:jpeg} shows the advantage of \textbf{IAP} to enhance the robustness against different corruptions. We think this is due to the fact that the message diffusion module (in Sec.\ref{decoder}) of \textbf{IAP} diffuses the message contained in adversarial examples to different frequency channels. Image corruptions may disrupt some frequency channels, and \textbf{IAP} can still extract messages from other channels.  

\subsection{Ablation Study and Further Discussion} \label{ablation}

\textbf{1$)$ Why not directly combine the adversarial perturbation and watermark?}

\textbf{IAP} uses an encoder to map the original image and the corresponding information to an adversarial example, which aims to provide two-tier protection for facial images. Then can we achieve both initiative and passive protection by directly combining existing SOTA adversarial attack and watermarking methods?
As shown in Tab.\ref{tab:add_experiment}, we choose the MBRS\cite{jia2021mbrs} as the robust watermark method and three different adversarial attack methods. "MBRS+I-FGSM" means that we first inject the watermark into the input image with MBRS and then use I-FGSM to generate the adversarial example for this image. "I-FGSM+MBRS" means that we first use I-FGSM to get the adversarial example for the input image and then inject the watermark into this adversarial example. Lower BER indicates higher extraction accuracy. This experiment shows that directly combining the adversarial perturbation and watermark can not extract the watermark correctly and slightly reduce the attack success rate. The BER for extracting the watermark generated by "MBRS+I-FGSM" from the image manipulated by HiSD is $38.39\%$, which is $37.29\%$ higher than our \textbf{IAP} in the same case. In addition, \textbf{IAP} generates adversarial examples containing information in one forward pass, which is convenient for use. 

\begin{table}[htb]
\caption{The combination of adversarial attack and watermark method. We show the average BER for extracting messages from the images manipulated by HiSD and StarGAN and the attack success rate on HiSD and StarGAN. PSNR is adjusted to about 36 by the perturbation strength factor $\beta$ for fair. The best results are in bold.} 
\centering
\scalebox{1.15}{
\begin{tabular}{c|cc|cc}
\hline \hline
\multicolumn{1}{c|}{\multirow{2}{*}{Methods}} & \multicolumn{2}{c|}{HiSD} & \multicolumn{2}{c}{StarGAN} \\ \cline{2-5} 
\multicolumn{1}{c|}{} & \multicolumn{1}{l|}{BER $\downarrow$} & ASR $\uparrow$& \multicolumn{1}{l|}{BER $\downarrow$} & ASR  $\uparrow$\\ \hline
MBRS+I-FGSM & \multicolumn{1}{l|}{38.39} & \textbf{1.0000} & \multicolumn{1}{l|}{40.62} & 1.0000 \\ 
MBRS+PG & \multicolumn{1}{l|}{18.92} & 0.9498 & \multicolumn{1}{l|}{14.08} & 0.9905 \\ 
MBRS+CMUA & \multicolumn{1}{l|}{29.62} & 0.9607 & \multicolumn{1}{l|}{23.51} & 0.9912 \\ \hline
I-FGSM+MBRS & \multicolumn{1}{l|}{38.28} & 1.0000 & \multicolumn{1}{l|}{39.56} & 1.0000 \\ 
PG+MBRS & \multicolumn{1}{l|}{16.75} & 0.9218 & \multicolumn{1}{l|}{12.95} & 0.9818 \\ 
CMUA+MBRS & \multicolumn{1}{l|}{27.78} & 0.9600 & \multicolumn{1}{l|}{15.92} & 0.9914 \\ \hline
IAP (ours) & \multicolumn{1}{l|}{\textbf{1.10}} & 0.9902 & \multicolumn{1}{l|}{\textbf{0.58}} & \textbf{1.0000} \\ \hline \hline
\end{tabular}\label{tab:add_experiment}}
\end{table}

\textbf{2$)$ How about using the LPIPS or MSE as the adversarial loss $\mathcal{L}_A$ in the IAP framework?}

In this paper, we propose FCM loss which extracts the facial features by a deep neural network and then compares these features using cosine similarity loss.
LPIPS \cite{zhang2018unreasonableLPIPS} is also a feature-level similarity measurement and works well in comparing the perceptual similarity of images. However, in facial protection tasks, we should focus on the attribute features of the faces rather than other features in the image. In Fig.\ref{img:attackloss}, we show that LPIPS can not measure the similarity of two images according to the change in facial features. Tab.\ref{tab:lpips_MSE} shows that using the LPIPS instead of the FCM loss in our \textbf{IAP} framework can not effectively disrupt the Fixed-point GAN and HiSD.  

As we analyzed in Sec.\ref{attack_facial}, the pixel-level measurement MSE is not very suitable in facial protection tasks in which the facial features deserve more attention than image pixels. As shown in Tab.\ref{tab:lpips_MSE}, using MSE loss as an alternative to our FCM loss in the IAP framework can successfully disrupt the AttentionGAN and the Fixed-point GAN but suffers a decline in performance when attacking the HiSD and the StarGAN. Using FCM loss in our \textbf{IAP} framework can perform initiative protection well against different facial manipulation systems.

\begin{table}[htb]
\caption{The attack success rate of using LPIPS or MSE loss as the adversarial loss in our IAP framework. PSNR is adjusted to about 36 by the perturbation strength factor $\beta$ for fair. The best results are in bold.}
\centering
\scalebox{1.05}{
\begin{tabular}{c|c|c|c|c|c}
\hline
\hline
Dataset & Methods & HiSD & StarGAN & AGAN & FGAN \\ \hline
\multirow{3}{*}{CelebA} & LPIPS & 0.8067 & 0.9875 & 0.9782 & 0.1733 \\ 
 & MSE & 0.8317 & 0.5686 & 0.9742 & \textbf{1.0000} \\ 
 & FCM(ours) & \textbf{0.9883} & \textbf{1.0000} & \textbf{0.9850} & 0.9984 \\\hline
\multirow{3}{*}{LFW} & LPIPS & 0.7982 & 1.0000 & 0.9771 & 0.1145 \\ 
 & MSE & 0.8342 & 0.5967 & 0.9712 & \textbf{0.9974} \\ 
 & FCM(ours) & \textbf{0.9900} & \textbf{1.0000} & \textbf{0.9833} & 0.9911 \\ \hline \hline
\end{tabular}\label{tab:lpips_MSE}}
\end{table}

\textbf{3$)$ Can IAP indicate the protection state of an image?}

 In the previous sections, we show that \textbf{IAP} can effectively extract information contained in adversarial examples and distort the output of facial manipulation systems. {Here we show that \textbf{IAP} can also indicate the protection state of an image because \textbf{IAP} can distinguish the natural images (without being protected by \textbf{IAP}) and the protected images with high accuracy.}

\begin{table}[htbp]
\caption{We show an example for the message $M_{in}$ contained in the IAP, the extracted message $M_{out}$, and the invalid message extracted from the images without protection. We illustrate seven bits of the message.}
\centering
\scalebox{0.82}{
\begin{tabular}{c|ccccccccc}
\hline
\hline
$M_{in}$ & 1 & 0 & $\cdots$ & 1 & 1 & 1 & $\cdots$ & 0 & 0 \\
$M_{out}$ & 1.035 & 0.018 & $\cdots$ & 1.081 & 1.078 & 1.022 & $\cdots$ & 0.007 & -0.027 \\
$M_{invalid}$ & 0.419 & 0.559 & $\cdots$ & 0.481 & 0.380 & 0.141 & $\cdots$ & 0.379 & 0.402\\
\hline
\hline
\end{tabular}\label{tab:message}}
\end{table}

 The input message $M_{in} \in {\{0,1\}}^L$ is either 0 or 1 at each bit. In practice, the bit in the extracted message $M_{out}$ is set to 1 if it is greater than 0.5; otherwise, it is set to 0. As shown in Tab.\ref{tab:message}, qualitatively, the message $M_{out}$ extracted from the protected images is close to 0 or 1 at each bit while the message extracted from the unprotected images $M_{invalid}$ is close to 0.5. 
Consequently, the value of the extracted message can be used to distinguish whether the image is protected by \textbf{IAP}.
 We refer to the mean distance between each bit of the message and 0.5 as the message score and use this message score to distinguish whether an image is protected by \textbf{IAP}. We note the message score as MS:
  \begin{equation}
  \begin{aligned}
    \mathop{{MS}} \triangleq \frac{1}{L}\sum\limits_{i=1}^L | m_i-0.5 | ,
 \label{MS}
 \end{aligned}
 \end{equation}
 where $m_i$ represents the i-th bit of the message.
 
\begin{figure}[htbp]
\centering
\includegraphics[width=0.486\textwidth]{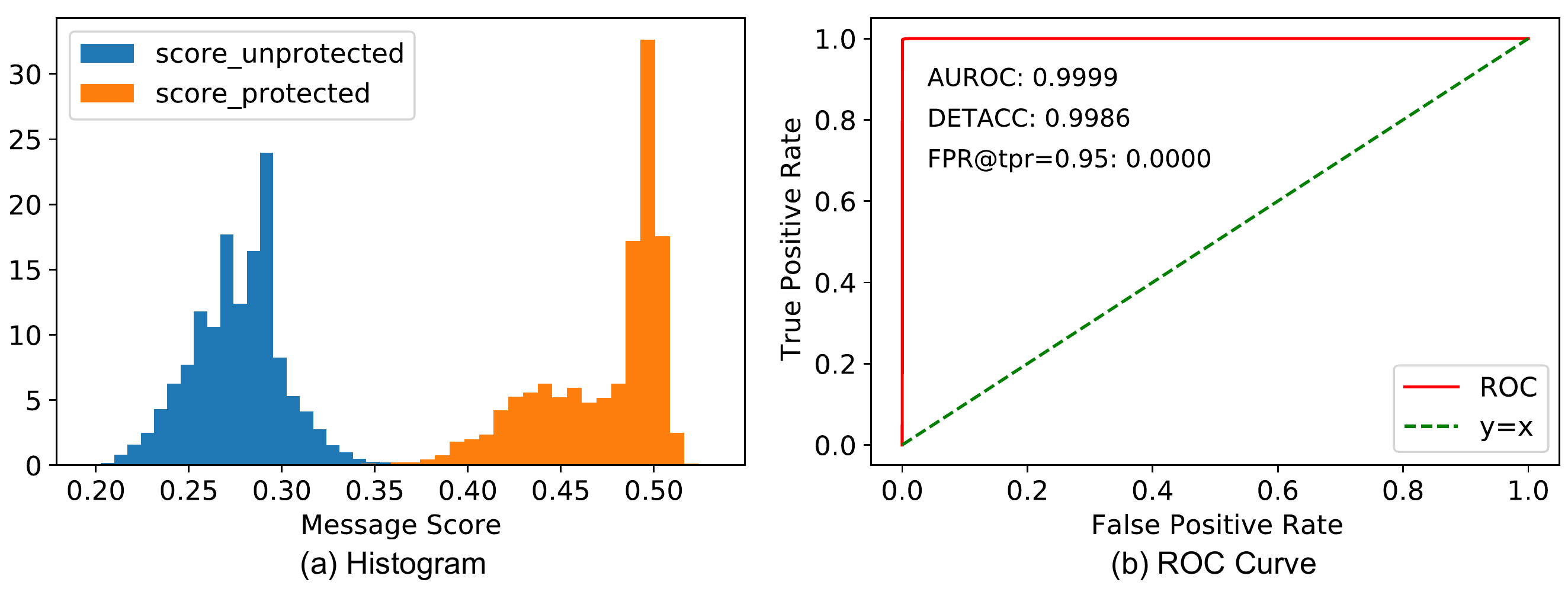}
\caption{(a) Frequency histogram of the message score on protected images and unprotected images. (b) The Receiver Operating Characteristic (ROC) curve for the message score on protected images and unprotected images. We refer to the protected images as the positive examples and the unprotected images as the negative examples. We show the AUROC, AUIN and FPR$@$tpr=0.95 in the upper left of the image.}
\label{img:message_score}
\end{figure}
 
 The quantitative results {are} shown in Fig.\ref{img:message_score}. We calculate the message score on 3000 unprotected images ("score$\_$unprotected" in Fig.\ref{img:message_score} (a)), including the original unprotected images and the unprotected images manipulated by different facial manipulation systems. We also calculate the message score on 3000 protected images ("score$\_$protected"), including the original protected images and their manipulated counterparts. 
 Fig.\ref{img:message_score} (a) intuitively shows the difference in the message score between protected images and unprotected images. We refer to the protected images as the positive examples and the unprotected images as the negative examples. Then we illustrate the Receiver Operating Characteristic (ROC) curve for the message score on protected images and unprotected images.
 The area under the receiver operating characteristic (AUROC) is a performance metric that is usually used to evaluate whether we can identify positive and negative examples. The AUROC is 99.99$\%$ which means that the protected images and the unprotected images can be well distinguished. "DETACC: 0.9986" means the detection accuracy at the best threshold obtained by the enumeration method is 99.86$\%$. "FPR$@$tpr=0.95: 0.0000" means that when the True Positive Rate (correctly regard the protected images as the protected images) is 95$\%$, the probability of False Positive Rate (wrongly regard some unprotected images as the protected images) is 0.00$\%$.
 Overall, our proposed \textbf{IAP} can successfully distinguish whether the image is protected that can be traced back to its source image in the database or the image is unprotected that can be added to the image database.
 
 {
 \textbf{4$)$ Can IAP protect the images against unknown systems?}
 
 Recent initiative defense methods \cite{ruiz2020disrupting,huang2021initiative,huang2021cmuawatermark} mainly focus on how to attack the white-box GANs because the transferability of the adversarial perturbation across different GANs can be very low. {Unlike} the common initiative defense methods that feel quite helpless when encountering unknown systems, our two-tier protection method \textbf{IAP} can still passively detect whether the image is fake when the initiative defense fails.
 
The previous subsections show that our IAP is robust against different corruptions and known manipulation systems. Here we show that IAP can also provide protection for images against systems for identity swap (SimSwap \cite{chen2020simswap}) and face reenactment ( Wav2Lip \cite{prajwal2020lip}). {The vulnerability between these two systems can be different from the facial manipulation systems used to train the adversarial perturbation generator, causing the failure of the initiative defense against these two systems.} The {standard adversarial} perturbation may feel quite helpless in this scenario. But our two-tier protection method IAP can still extract the identity message from the images correctly to distinguish whether the image is modified and stop the spread of fake images. In Fig. \ref{img2:simswap}, we show that SimSwap can escape the initiative protection provided by IAP and successfully swap the identity of the image. However, IAP can still extract the identity message of the image correctly, contributing to provenance tracking and detecting the fake image. In Fig. \ref{img2:wav2lip}, we use Wav2Lip to generate a video with the unexpected voice from a protected video (inject perturbation on the frames of the video). IAP can correctly recover the identity message from the modified video and distinguish the video whether a fake video by comparing it with the source video. Similar to the barcode or QR code used to trace the source and ensure the safety of the commodities, our IAP can be used to trace the source of the facial image and protect the image spread on social media platforms.

\begin{figure}[htbp]
\centering
\includegraphics[width=0.49\textwidth]{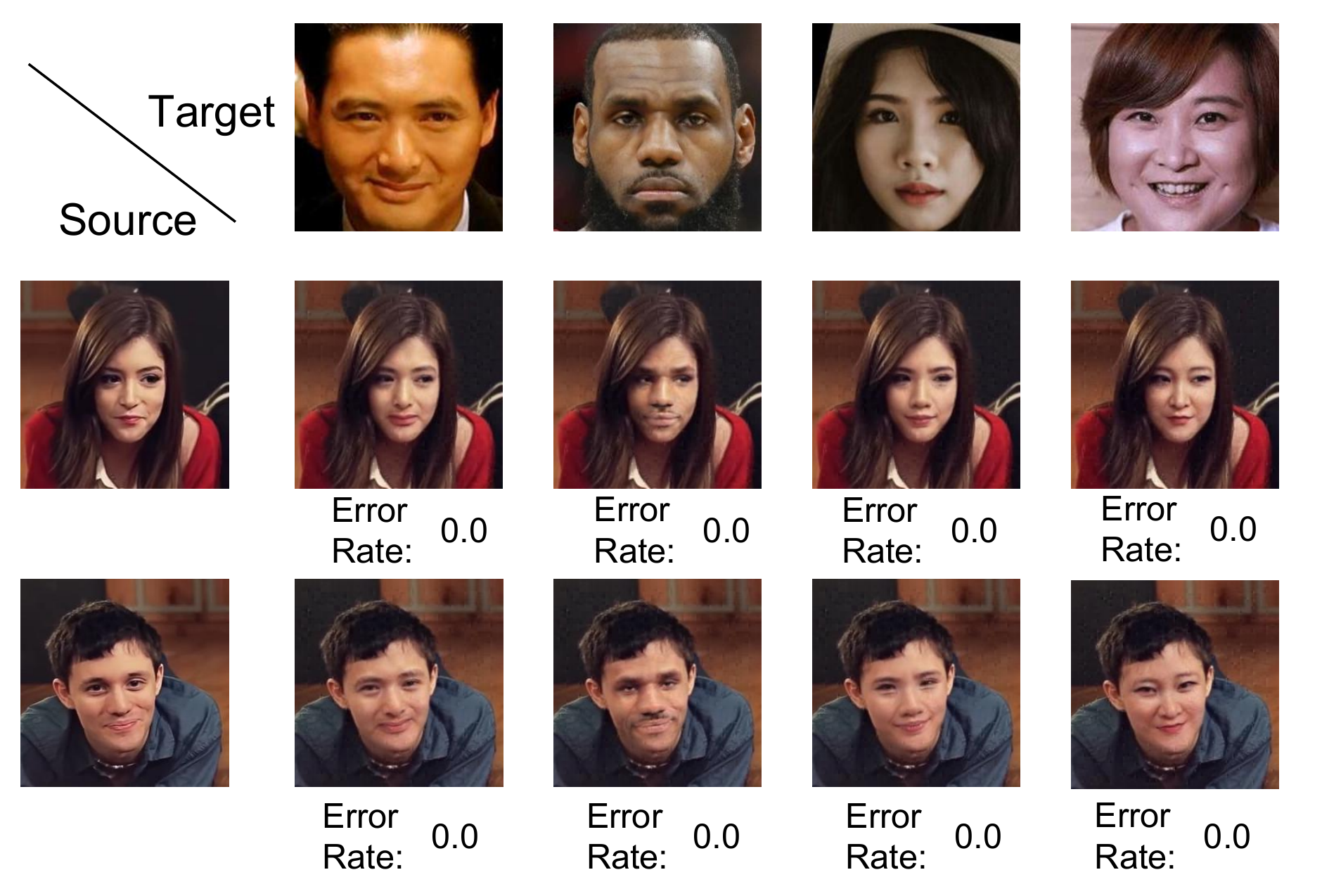}
\caption{SimSwap \cite{chen2020simswap} can swap the identity of the protected image successfully, but we can recover the message in the protected image with a bit error rate is 0. That is to say, we can successfully perform provenance tracking and stop the spread of fake images.}
\label{img2:simswap}
\end{figure}

\begin{figure}[htbp]
\centering
\includegraphics[width=0.49\textwidth]{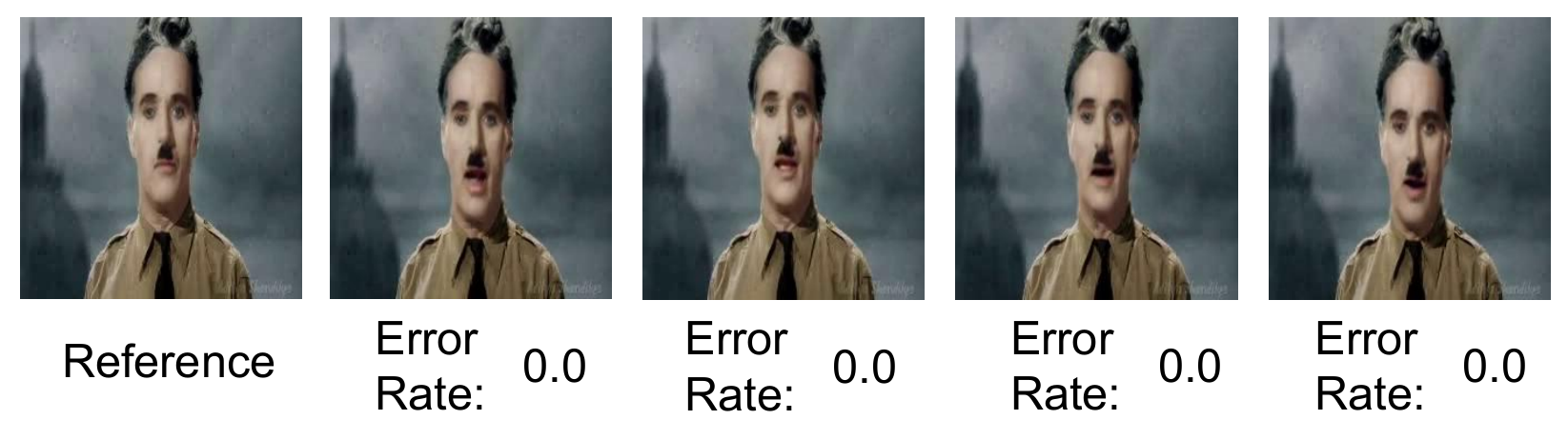}
\caption{Wav2Lip \cite{prajwal2020lip} can perform the reenactment on the protected image successfully, but we can also recover the message in the protected image with a bit error rate is 0. That is to say, we can successfully perform provenance tracking and stop the spread of fake images.}
\label{img2:wav2lip}
\end{figure}

These experiments demonstrate the effectiveness and importance of our two-tier protection method IAP. {Standard adversarial} attacks can utilize the vulnerability of the facial manipulation systems to perform initiative defense but may fail to protect the image when encountering an unknown system. Our IAP overcomes the shortcomings of existing initiative defense methods by provenance tracking. 
}

\section{Conclusion}
The privacy-preserving is an essential issue.
In this paper, we propose a novel two-tier facial protection method named Information-containing Adversarial Perturbation (\textbf{IAP}).
For one thing, \textbf{IAP} can work as an initiative defense method by disrupting the performance of the facial manipulation systems, which prevents the facial image from malicious manipulation. For the other thing, the information contained in \textbf{IAP} can be used as a clue for provenance tracking, which is helpful for fake image detection or other passive methods. To the best of our knowledge, our \textbf{IAP} is the first work that incorporates the advantages of initiative protection and passive protection.
Further, we propose the feature-level correlation similarity for comparing the facial images, which is more suitable for facial protection than existing methods. Moreover, we design a spectral diffusion module to improve the robustness of the information contained in the adversarial example against facial manipulation systems and other image corruptions. 
Extensive experiments demonstrate that our proposed \textbf{IAP} can effectively disrupt the facial manipulation systems and the information contained in the adversarial example can be correctly extracted from both the adversarial examples and their manipulated counterparts.
Lastly, the merit of embedding meaningful information into the adversarial perturbation and fusing the essence of initiative and passive protection is not limited to facial protection. It may potentially improve privacy protection in a broader aspect as well.

\section*{Acknowledgments}
This work was supported in part by the National Key Research and Development Program of China under Grunt No. 2020AAA0140000, and by the Fundamental Research Funds for the Central Universities, and by Alibaba Group through Alibaba Research Intern Program.

 {\small
\bibliographystyle{IEEEtranN}

\bibliography{newIEEEabrv}
}

\vfill

\end{document}